\documentclass[letterpaper, 10 pt, conference]{ieeeconf}  

\IEEEoverridecommandlockouts                              

\usepackage{cite}
\usepackage{amsmath,amssymb,amsfonts}
\usepackage{algorithmic}
\usepackage{graphicx}
\usepackage{subcaption}
\usepackage{textcomp}
\usepackage{xcolor}
\usepackage{tikz} 
\usepackage{tikzscale}
\usepackage{pgfplots}
\usepackage{pgfgantt}
\usepackage{pdflscape}
\pgfplotsset{compat=newest} 

\newlength\figW
\newlength\figH
\usetikzlibrary{arrows,decorations.markings}

\title{\LARGE \bf
Priority Maps for Surveillance and Intervention of Wildfires and other Spreading Processes }

\author{Vera L. J. Somers and Ian R. Manchester
\thanks{The authors are with the Australian Centre for Field Robotics (ACFR), School of Aerospace, Mechanical and Mechatronic Engineering,
        University of Sydney, NSW 2006, Australia
        {\tt\small \{v.somers, i.manchester\}@acfr.usyd.edu.au}}%
}

\begin{document}

\maketitle
\thispagestyle{empty}
\pagestyle{empty}

\begin{abstract}
Unmanned Aerial Vehicle (UAV) path planning algorithms often assume a knowledge reward function or priority map, indicating the most important areas to visit. In this paper we propose a method to create priority maps for monitoring or intervention of dynamic spreading processes such as wildfires. The presented optimization framework utilizes the properties of positive systems, in particular the separable structure of value (cost-to-go) functions, to provide scalable algorithms for surveillance and intervention. We present results obtained for a 16 and 1000 node example and convey how the priority map responds to changes in the dynamics of the system. The larger example of 1000 nodes, representing a fictional landscape, shows how the method can integrate bushfire spreading dynamics, landscape and wind conditions. Finally, we give an example of combining the proposed method with a travelling salesman problem for UAV path planning for wildfire intervention.
\end{abstract}

\section{INTRODUCTION}

The application of Unmanned Aerial Vehicles (UAVs) for disaster response is an active area of research \cite{Messinger2016,Mcclean2010,AmericanRedCross2015}. In the case of wildfires, a.k.a. bushfires, there is an interest in exploring how UAVs can offer quicker and more efficient situational awareness as well as intervention via waterbombing. Their mobility, aerial view and ability to quickly gather and transmit data offer the potential of making them a central tool in future fire assessment and response \cite{Howden2008,Nigam2014}. Although many fire propagation models and simulators exist (e.g. \cite{Karafyllidis1997a,  Johnston2006}) and path planning for spatial monitoring is well advanced (e.g. \cite{yu2016correlated, smith2012persistent, penicka2017dubins}), the link connecting them is still missing. Other spreading processes such as flood and the spread of disease pose similar challenges \cite{Nowzari2015}. In this paper, we aim to bridge this gap by generating \textit{priority maps} for robotic surveillance or intervention that depend on the dynamics of the spreading process.

Pioneering work in firefront propagation modelling was done by Rothermel in \cite{Rothermel1972,Rothermel1983New} and almost all available fire models build upon this approach. Fire spread models can be split out in two categories regarding their landscape representation: continuous and discrete, of which discrete models are better in dealing with heterogeneous data. In the discrete landscape category, cellular automata (CA) models, which are based on fire transition probabilities from one cell to another, are most commonly used due to their ease of implementation, low computational cost and suitability for heterogenous conditions \cite{Sullivan2009}. When based on accurate information about vegetation, terrain, and weather conditions, CA models have been shown to give quantitatively accurate predictions when compared to real wildfire spread \cite{Alexandridis2008a,Kelso2015}. However, a downside of CA models is that they are not directly amenable to optimization-based analysis and decision-making. 

 A closely related process is the outbreak and spread of epidemics, for which CA-type models are usually simplified to ordinary differential equation (ODE) models such as the spreading Susceptible-Infected-Susceptible (SIS) model \cite{kermark1927contributions,bailey1975mathematical}. A linear approximation of the nonlinear model is given in \cite{Ahn2013,VanMieghem:2009,Nowzari2016}. As proven by \cite{VanMieghem:2009} the linear model can be considered an upperbound for the non-linear model and can therefore be seen as a worst-case scenario. 

A priority map defines critical areas, or nodes on a graph, from which the the spread of a fire (or flood, or disease) would be particularly catastrophic in terms of risk of life and property. In epidemics, many different methods for ranking nodes exist, where the most common ones are based on graph topology, such as degree \cite{Madar2004}, reproduction number \cite{Nowzari2016} or eigenvector centrality\cite{youssef2011individual}. However, in \cite{ Liu2016} is shown that considering topological features alone is in most cases not enough to indicate influential nodes. In wildfire models, it is particularly important to consider spatially-varying spreading rates, due to the effects of vegetation, terrain, and weather. These are usually not considered in epidemic models.

Once the critical nodes are defined, an interesting problem is to look at strategies to influence the system and reduce the impact of the spreading process within a budgeted constraint, i.e. develop a cost-to-go map indicating which nodes to `attack'. However the problem of how many and which nodes or links to remove turns out to be respectively NP complete and NP hard \cite{van2011decreasing}. Therefore a more suitable approach is to reduce the impact of a node by tuning and controlling the spreading rate, as demonstrated by \cite{Preciado}. Solutions to the optimal control problem for vaccination and patching for regulating infection levels are presented in \cite{Giamberardino2017,Bloem2008,Khanafer}.

In \cite{Lindmark} parameters for controllability of a network are identified, where the main result is based on topological features. Similar in \cite{Dhingra2018}, leader selection and its influence on stability and optimal performance in directed graphs is explored. Furthermore, combination drug therapy is studied by the authors and solved by influencing the main diagonal of the dynamic matrix. Another method for sparse resource allocation for linear network spread dynamics by modifying the diagonal is given in \cite{Torres2017}. The dominant eigenvalue of the dynamics is minimized to achieve this. 

The main idea of this paper is to determine priority maps for both surveillance and intervention scenarios by considering a bushfire as a spreading process on a graph with similar dynamics as other wide modeled network spreading processes such as epidemics and computer viruses, e.g.  as given in \cite{Bloem2008,Torres2017}. Because the spreading rate will always be nonnegative, it falls within the category of positive systems \cite{berman1994nonnegative}. Such systems allow for control \cite{briat2013robust, rantzer2015scalable} and identification \cite{umenberger2016scalable} algorithms based on linear programming, with significantly improved scalability compared to general linear systems. 

In this paper we present a method to prioritize nodes in a network for different scenarios for positive systems with an application to UAV path planning. The presented method distinguishes itself on two main points from previous research: 1) it is not topological dependent and therefore, can deal with network structures, such as grids, where all nodes have similar degrees and 2) it also considers nondiagonal entries for control resources and is therefore more flexible and suitable for different types of spreading dynamics. We explain this further in Section II, where we give a framework for modeling spreading processes of positive systems on a network and develop a method to obtain priority maps for two different scenarios: the surveillance and intervention problem. The first is explained further in Section III, whereas the latter is evaluated in Section IV. With examples we convey that the proposed method is easy applicable and provides quick, good and solid results. 

\section{PROBLEM AND MODEL FORMULATION}
Cellular automata bushfire simulations like \cite{Karafyllidis1997a,Johnston2006,Alexandridis2008a} use an underlying stochastic model with Markov transition probabilities. Let us consider a graph with $n$ nodes, where each node $i \in \{1, 2, ..., n\}$ has a state $x_i(t)$ associated with it. This state now transitions following a set of rules. If a node $i$ is on fire, $x_i(t)=1$, it recovers with probablity $\delta_{i}\Delta t$ to $x_i(t+\Delta t)=0$ and the fire spreads to a neighbouring node with probability $\beta_{i} \Delta t$.

Using a mean-field estimation and Kolmogorov forward equations the deterministic model can be approximated from this stochastic model \cite{Nowzari2015,Preciado2014}. Here, the main assumption taken is that all the random variables have zero covariance. It is proven in \cite{Mieghem2011,Li2012} that the obtained approximated probabilities upper bound the actual ones, which has a positive effect on controlling the underlying stochastic process.

\begin{figure}[!ht]
\centering
\begin{subfigure}[b]{0.32\linewidth}
        \includegraphics[width=1\linewidth]{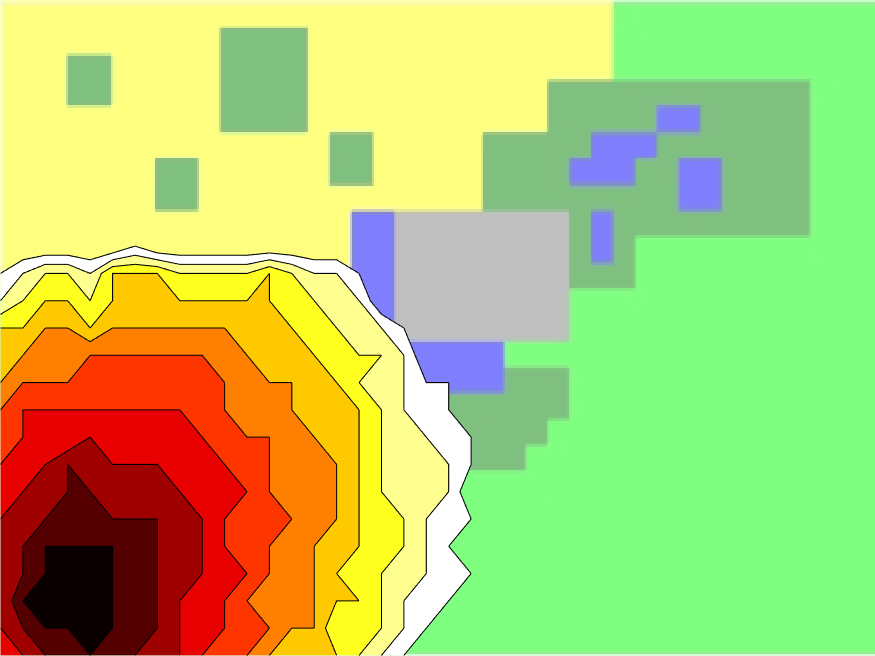}
      \caption{Cellular Automata}
      \label{fig:SpreadCA}
  \end{subfigure}
     \begin{subfigure}[b]{0.32\linewidth}
    \includegraphics[width=1\linewidth]{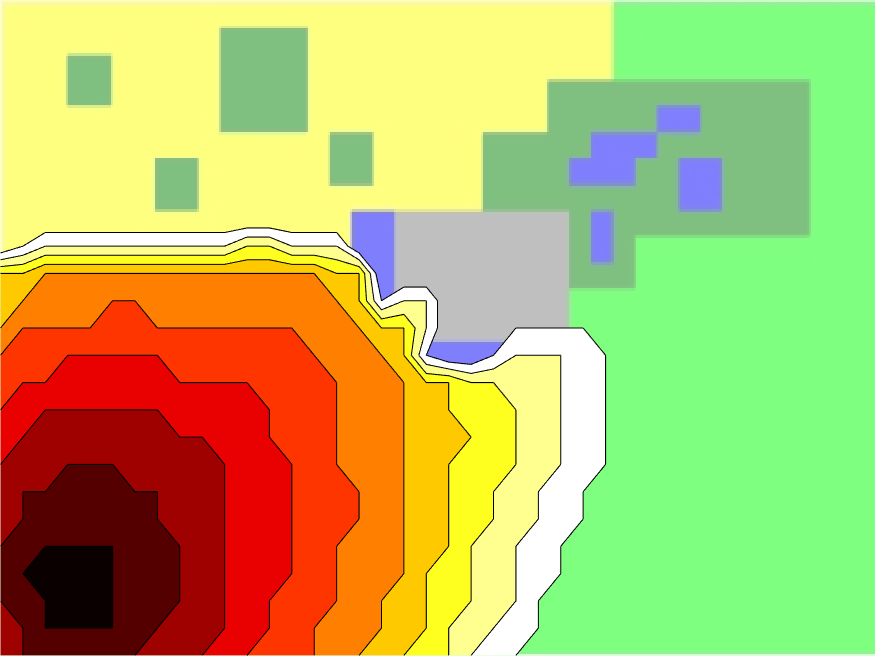}
      \caption{Non Linear}
      \label{fig:SpreadNonLin}
  \end{subfigure}
     \begin{subfigure}[b]{0.32\linewidth}
    \includegraphics[width=1\linewidth]{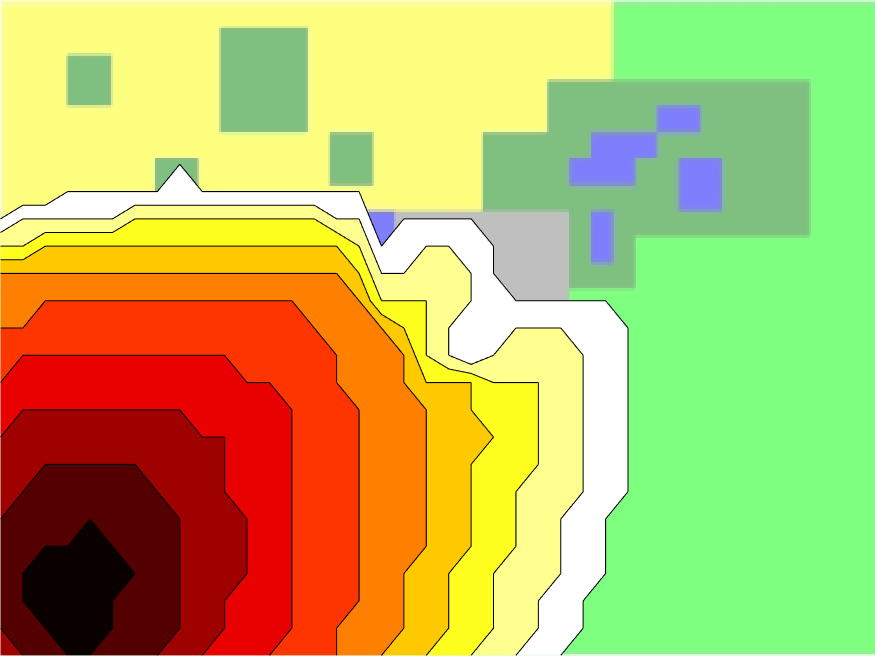}
      \caption{Linear}
      \label{fig:SpreadLin}
  \end{subfigure}
      \caption{Fire spread comparison of the different models on a fictional landscape. The different colors indicate spreading over time, where lighter indicates further in time.}
\label{fig:Spread}
\end{figure}

The linear model is finally obtained by linearizing this deterministic model around the fire-free equilibrium point ($\mathbf{x}=0$) \cite{Preciado2014}. This model again upper bounds the nonlinear model \cite{VanMieghem:2009,Preciado}. Therefore the probabilities determined can be seen as the worst case scenario as shown in Fig. \ref{fig:Spread}. Here it can be seen that the linear model realistically upperbounds the non linear deterministic one, which upperbounds the stochastic model. Closer to the equilibrium point the linear model more closely approximates the stochastic process whereas over time the error accumulates.

We can now define the system on the graph by the linear differential equation:
\begin{equation}
\label{eq1}
   \dot{x}(t)=Ax(t) 
\end{equation}
where $x(t)=[x_{1}(t),...,x_{n}(t)]^{T}$ with $t\geq0$, is the state of the system and the sparse state matrix $A$ is defined by the linearized spreading dynamics 
 \begin{equation}
  \label{epi}
 a_{ij} = 
\begin{cases}
-\delta_{i}  \le 0 &\quad \text{if}\quad  i=j, \\
\beta_{ji} \ge 0&\quad \text{if}\quad  i\neq j, (i,j) \in E, \\
0  &\quad \text{otherwise.}
 \end {cases}
 \end{equation}
Here $E$ is the set of edges of the graph and therefore the off-diagonal entries $a_{ij}$ are determined by the spreading rate $\beta_{ij}$ and the adjacency matrix of the graph, whereas the diagonal entries are given by $-\delta_i$, the recovery rate. 
 Because all off-diagonal entries $a_{ij}$ are assumed to be nonnegative, $A$ is Metzler and the system is positive, i.e. if $x_i(0)\ge 0$ for all $i$, then $x_i(t)\ge 0$ for all $t\ge 0$ \cite{berman1994nonnegative}.
  
The cost function $J$ associated with the system is defined as  
\begin{equation}
\label{eq:cost2}
J=  \int_{0}^\infty e^{-rt} Cx(t)dt
\end{equation}
where $C= [c_1, ..., c_n]$ is a row vector defining the cost associated with each node $i$, with each $c_i\ge 0$. For example, in a wildfire the cost of a populated area burning is much higher than open grassland. The discount rate $r>0$ can be tuned to emphasize near-term cost over long-term cost. In surveillance problems discounting is necessary for the above integral to exist since the spreading process dynamics of interest are generally unstable.

The question now remains how to define the priority $p$ of the nodes in the graph. In this paper we use the property of positive systems that they
 have \textit{linear} and hence \textit{separable} Lyapunov and storage functions, which has previously been used for scalable methods of robust control and identification \cite{rantzer2015scalable, briat2013robust, umenberger2016scalable}. We note that this property also extends to certain classes of nonlinear systems \cite{dirr2015separable, manchester2017existence}.
 
\subsection{Surveillance Problem}
\label{subsec:SP}
In the surveillance problem, the dynamics \eqref{eq1} are fixed and the aim is to identify the future cost of a fire spreading from each node. This then serves as a priority for surveillance. Using the properties of positive systems, we propose to find a priority vector $p$ such that
\begin{equation}
\label{eq:R2}
   \int_{0}^\infty e^{-rt} Cx(t)dt = \sum_{i=1}^n p_i x_i(0)
\end{equation}
where $c_{i}$ can be seen as the cost of each node burning, and $p_{i}x_i(0)$ as the discounted cost-to-go associated with each node $i$. The important point is the separable structure of this cost bound, i.e. it is the sum of terms each dependent on a single node $i$ without cross-terms, and hence these terms can serve as priority indicators for the nodes. In this paper, we focus on using $p_i$ as the priority for node $i$, but if state estimates are available then $p_{i}x_i(0)$ can also be used.

For any (unstable) matrix $A$, there exists a discount rate $r>0$ sufficiently large that $A-rI$ is Hurwitz-stable, i.e. all eigenvalues have negative real parts. Under this assumption, the node priorities can be calculated via the simple formula:
\begin{equation}
\label{eq:MM}
p^{T}=C(rI-A)^{-1}.
\end{equation}
The non-negativity of each $p_i$ follows from the fact that $rI-A$ is a positive-stable M-matrix, and hence has an elementwise non-negative inverse \cite[p.~134]{berman1994nonnegative}, i.e. all elements of the matrix $(rI-A)^{-1}$ are non-negative, as are all elements of $C$ by construction. Then \eqref{eq:MM} implies \eqref{eq:R2} since integrating over the time interval $[0,\infty)$ both sides of
\[
\frac{{d}}{{dt}} \left (e^{-rt}p^{T}x(t) \right )= - e^{-rt} Cx(t), 
\]
 gives \eqref{eq:R2}, while expanding the left-hand-side gives
\begin{equation*}
\begin{split}
e^{-rt}p^{T}\dot{x}(t) - re^{-rt}p^{T}x(t) & = - e^{-rt} Cx(t), \\
p^{T}(rI-A)x(t)&= Cx(t).
\end{split}
\end{equation*}
which must hold for all $x(t)$ hence implies \eqref{eq:MM}.

While \eqref{eq:MM} is suitable for calculating surveillance priorities, we also provide an equivalent formulation as a linear program (LP) which is suitable for extension to intervention problems. The equivalent LP is
\begin{equation*}
\begin{split}
    \text{minimize} & \quad \quad  |p|_{1} \\
    \text{such that} & \quad \quad p \geq 0, \quad \quad p^{T}A - r p^{T} \leq - C
\end{split}
\end{equation*}
To show equivalence, let $\bar{p}$ be the priority vector calculated using (\ref{eq:MM}), then $\bar p\ge 0$ and $p^{T}A - r p^{T} =- C $ hence $\bar{p}$ is feasible for the LP, but any other feasible $p$ has $p^{T}A - r p^{T} \le - C$ and hence (via inverse-positivity) $p^T\ge =C(rI-A)^{-1} = p$, so 
$
|p|_{1}=\sum p_{i} \geq \sum \bar{p}_{i} = |\bar{p}_{1}|
$
and so  $p=\bar{p}$ is optimal for the LP.
 
\subsection{Intervention Problem}
\label{subsec:IP}
In the intervention problem, we consider the scenario where we assume that the system dynamics can be altered to obtain a reduced cost-to-go vector $p$ and the new closed loop system is given by (\ref{eq2}). The task is to reduce the discounted cost-to-go by finding a sparse set of locations for intervention within a budgeted constraint. Assuming a linear effect of intervention, the closed-loop dynamics become
\begin{equation}
\label{eq2}
   \dot{x}(t)=(A-K)x(t) 
\end{equation}
where $K$ is the control matrix and restricted to the total resource constraint $\sum k_{ij} \leq \Gamma$, where $\Gamma$ is the maximum budget available. Note that this imposes an $l^1$-type constraint on the elements of $K$ and hence encourages sparsity. A key difference here compared to \cite{Dhingra2018} and \cite{Torres2017} is that the control resources can be put on the nondiagonal entries of $K$, i.e. $k_{ij} \geq 0$ if $(i,j) \in E$ and $k_{ij} = 0$ otherwise (instead of $k_{ij} \geq 0$ if $i=j$). Hence, the spreading rate is influenced instead of the recovery rate. 

Extending the surveillance LP to this scenario we have
\begin{equation*}
\begin{split}
    \text{minimize} & \quad \quad  |p|_{1} \\
    \text{such that} & \quad \quad p \geq 0,\quad p^{T}A - r p^{T} - p^{T}K \leq - C \\
    & \quad \quad \sum_{ij} K \leq \Gamma, \quad K \geq 0, \quad  K \leq A\\
    & \quad \quad k_{ij}=0  \quad \forall (i,j) \notin E 
\end{split}
\end{equation*}
However the multiplication of decision variables $p^{T}K$ means this is not a linear program. Therefore following \cite{rantzer2015scalable} a new variable $q = p^{T}K$ is introduced, resulting in the updated constraint $p^{T}A - r p^{T} - q \leq - C$. However to take into account the restriction imposed upon the control matrix $K$ by the total resource constraint $\Gamma$ and the adjacency matrix, we take $P=\text{diag}(p)$ and $Q=PK$ to preserve the specified structure of K. This implies $\vec{1}P=p^{T}$ and $\vec{1}Q=q$, where $\vec{1}$ is the row vector with all ones of the appropriate dimension and the problem can be reformulated as
\begin{equation*}
\begin{split}
    \text{minimize} & \quad \quad  \sum_{i}^{n} P_{ii} \\
    \text{such that} & \quad \quad P \geq 0,\quad Q \geq 0 \\
    & \quad \quad \vec{1}(PA - rP - Q) \leq - C \\
     & \quad \quad \sum_{ij} P_{0}^{-1}Q \leq \Gamma, \quad P_{0}^{-1}Q \leq A \\
     & \quad \quad q_{ij}=0  \quad \forall (i,j) \notin E  
\end{split}
\end{equation*}
where $\sum_{ij} |P_{0}^{-1}Q| \approx \sum_{ij} |K_{ij}| \leq \Gamma $ and $P_{0}$ is the initial guess of P taken as the solution of the surveillance problem. The optimization problem now consists of minimizing the trace of $P$, while updating $P_{0}$ to the obtained $P$ till the outcome converges, i.e. $P_{0}=P$. 

Afterwards, the final control matrix $K=P^{-1}Q$. The optimization can be extended by including nodes as possible control resource options by changing the constraints on $k_{ij}$. 

\section{SURVEILLANCE RESULTS}
We look at two examples with different dynamics to evaluate the method presented in Section \ref{subsec:SP} and its response to changes in the different parameters.  

\subsection{Example 1: 16-node grid}
\begin{figure}[!ht]
\centering
\begin{subfigure}[b]{0.4\linewidth}
        \includegraphics[width=1\linewidth]{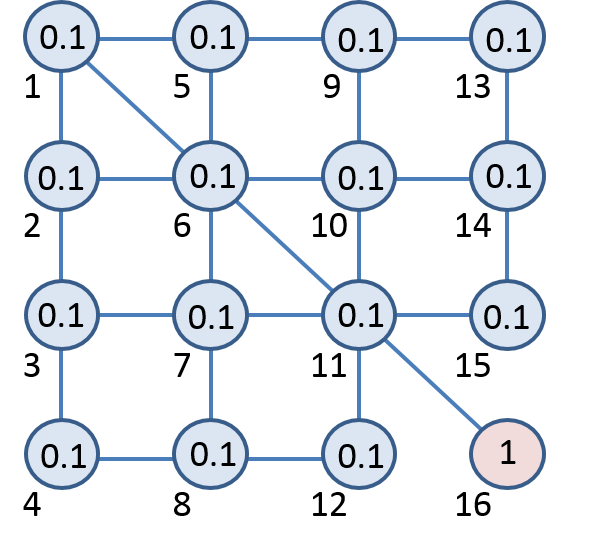}
      \caption{Graph}
      \label{fig:Graph1}
  \end{subfigure}
  ~ 
     \begin{subfigure}[b]{0.48\linewidth}
    \includegraphics[width=1\linewidth]{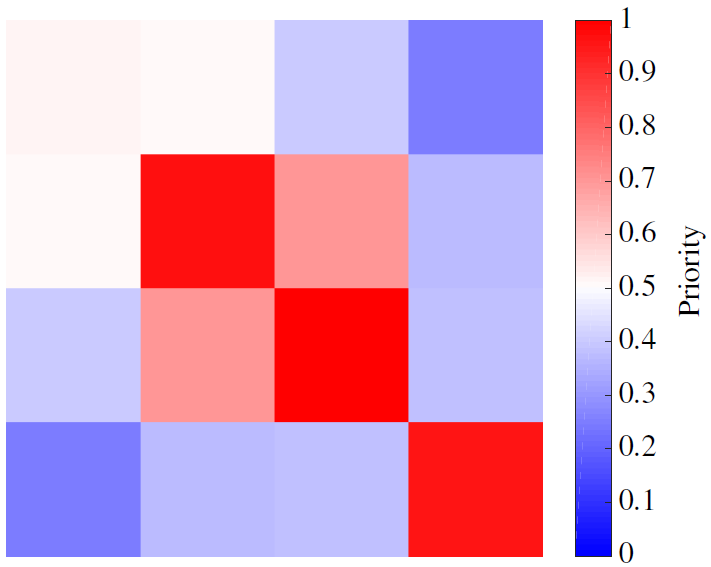}
      \caption{Normalized priority map}
      \label{fig:SPri}
  \end{subfigure}
      \caption{Graph and belonging priority map for a network with $n=16$ nodes with equal spreading rate of $\beta=0.5$ and recovery rate $\delta=0.2$ for all nodes. The cost $c_{i}= 0.1$ for $i=1,..,15$ and $c_{i}=1$ for $i=16$.}
\label{fig:Fig12}
\end{figure}

Let us consider a graph with $n=16$ nodes, connected as visualized in Fig. \ref{fig:Graph1} with a spreading rate $\beta = 0.5$ and recovery rate $\delta = 0.2$, where the cost $c_{i}=1$ for $i=16$ and $c_{i}=0.1$ for every other node. This results in a system for which $a_{ij}=-0.2$ if $i=j$, $a_{ij}=0.5$ if $(i,j)\in E$ and $a_{ij}=0$ otherwise. This can be seen as a landscape with a city, which has a high cost of burning down, on node $16$, which is only connected to node $11$. The objective is stated in Section \ref{subsec:SP} and taking $r=2$ the normalized results are shown in Fig. \ref{fig:SPri}. 

%

As expected due to the connectivity, equal spreading and recovery rate and with one high cost node, Fig. \ref{fig:SPri} shows high priorities for the diagonal nodes $6$ and $11$ leading up to the high cost node $16$.

\subsection{Example 2: 1000 node fictional landscape}
\begin{figure}[!ht]
  \centering
    \includegraphics[width=0.9\linewidth]{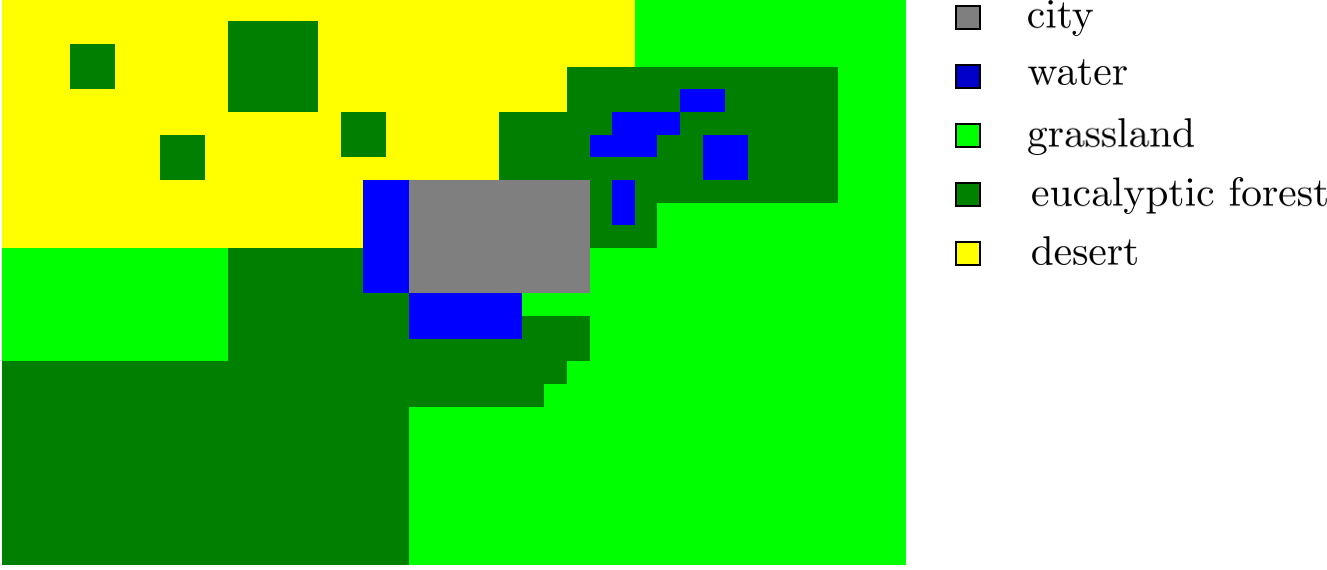}
    \caption{Fictional landscape with different area types, represented as a grid with $n=1000$ nodes.}
    \label{fig:Ex2}
\end{figure}

A more interesting problem is to look at the effect of different spreading rates $\beta$, which would be a more realistic representation of actual spreading processes. Therefore we consider the fictional landscape given in Fig. \ref{fig:Ex2} consisting of three different vegetation types, a city and water. It can be represented as a network graph with $n=1000$ nodes. The adjacency matrix and therefore, the set of Edges, is based on a 8 node spreading direction grid, i.e., the landscape is seen as a grid where each node is connected to its direct horizontal, vertical and diagonal neighbors. The recovery rate $\delta=0.2$ for all nodes, whereas $\beta$ is generated using data from cellular automata bushfire models presented in \cite{Karafyllidis1997a} and \cite{Alexandridis2008a}, where the bushfire spreading probabilities are determined based on real bushfire observations. For the spreading dynamics the baseline $\beta=0.5$, where a correction factor for the vegetation is taken as $\beta_{veg}=0.4$, $\beta_{veg}=1$  and $\beta_{veg}=1.4$ for respectively desert, grassland and eucalyptus forest. For the city this is $\beta_{veg}=0.5$ and for the unburnable water areas $\beta_{veg}=0$ is taken.  Finally, $\beta$ is corrected for spreading between diagonally connected nodes, following \cite{Karafyllidis1997a}. The cost of the city nodes is again $c_{i}=1$, wheras $c_{i}=0.01$ for all other nodes.

\begin{figure}
\centering
    \begin{subfigure}[b]{0.48\linewidth}            
             \def\svgwidth{0.99\linewidth}
              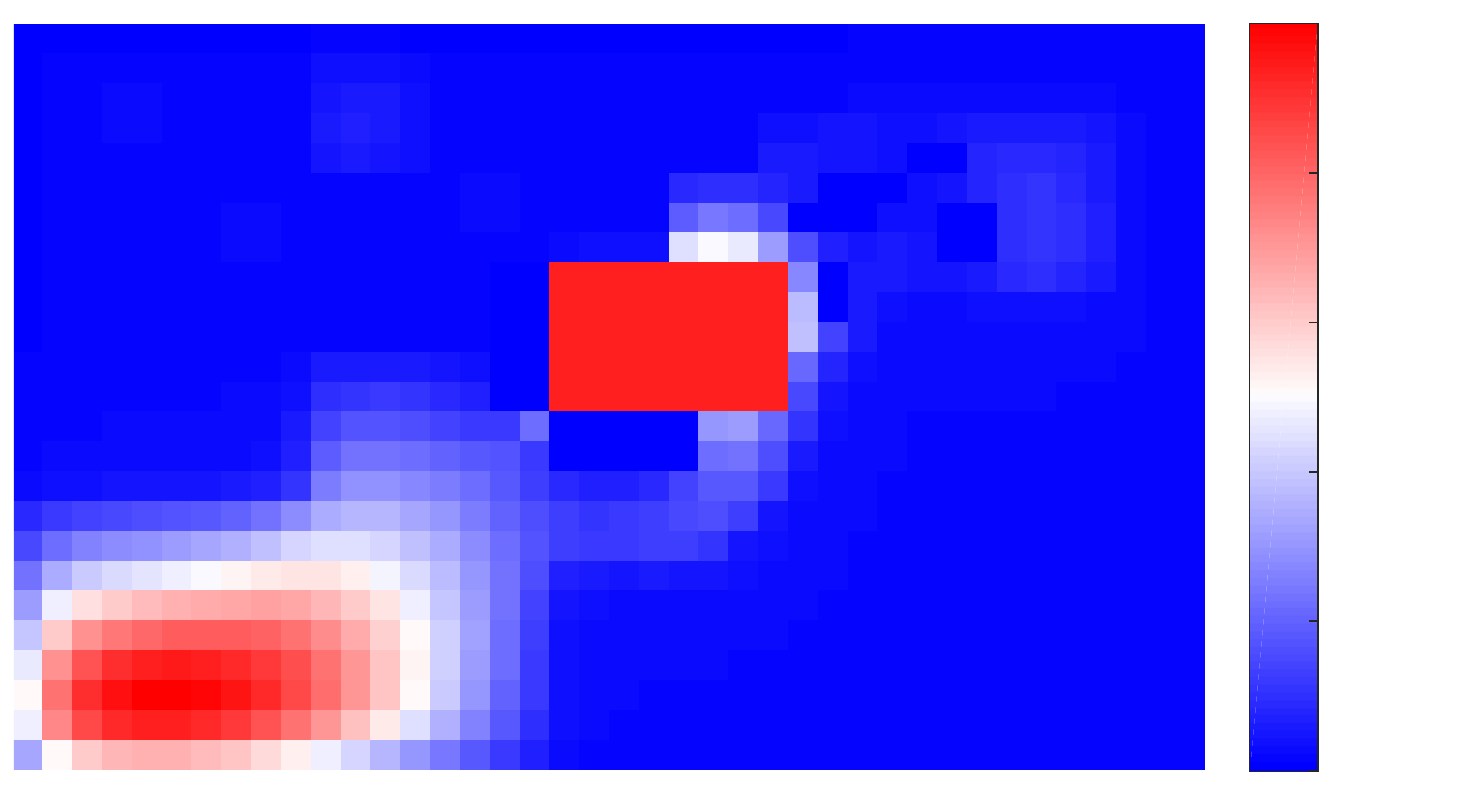
            \caption{Southwest}
            \label{fig:45}
    \end{subfigure}%
    \begin{subfigure}[b]{0.48\linewidth}
            \centering
           \def\svgwidth{0.99\linewidth}
              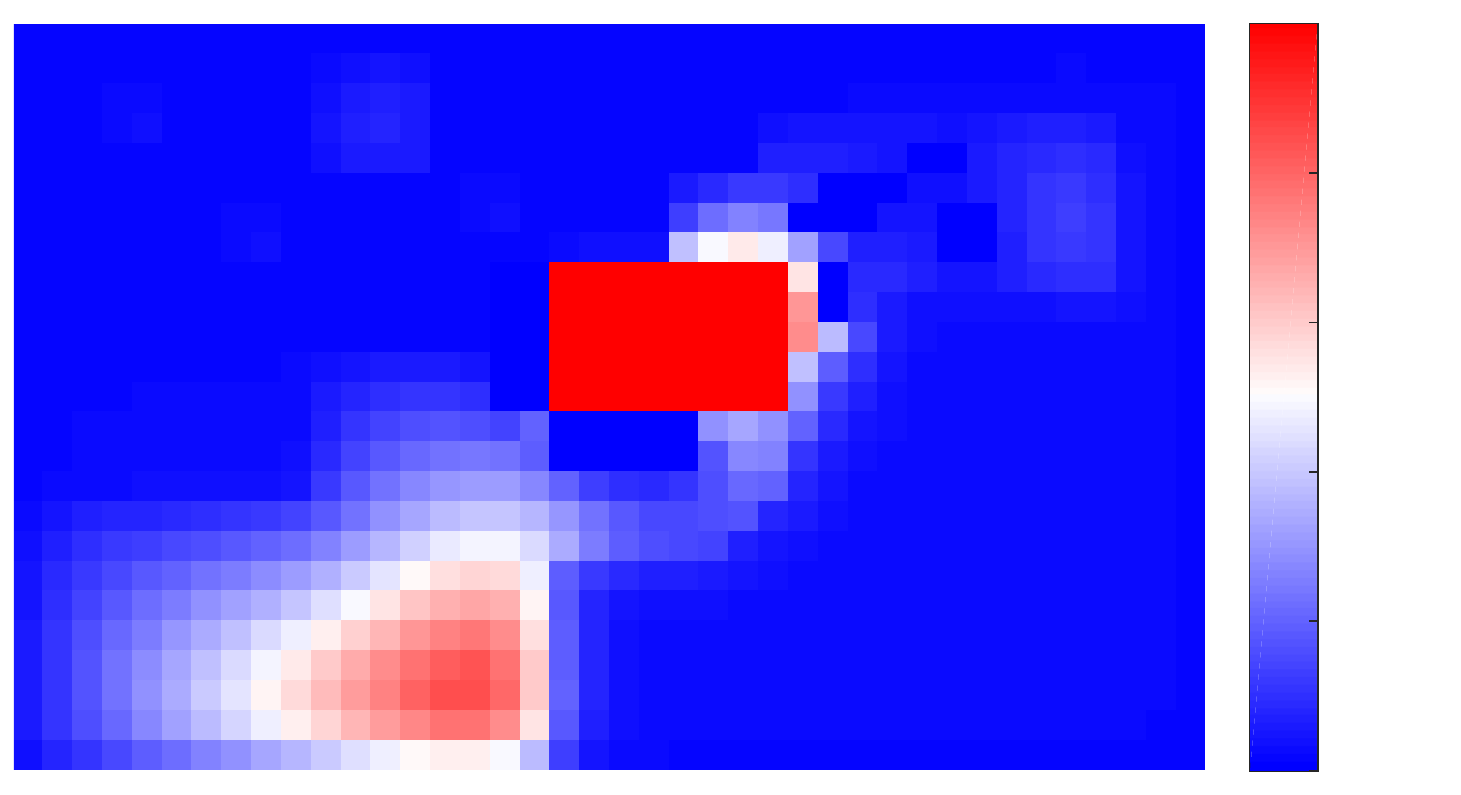
            \caption{Southeast}
            \label{fig:135}
    \end{subfigure}
    \begin{subfigure}[b]{0.48\linewidth}            
            \def\svgwidth{0.99\linewidth}
              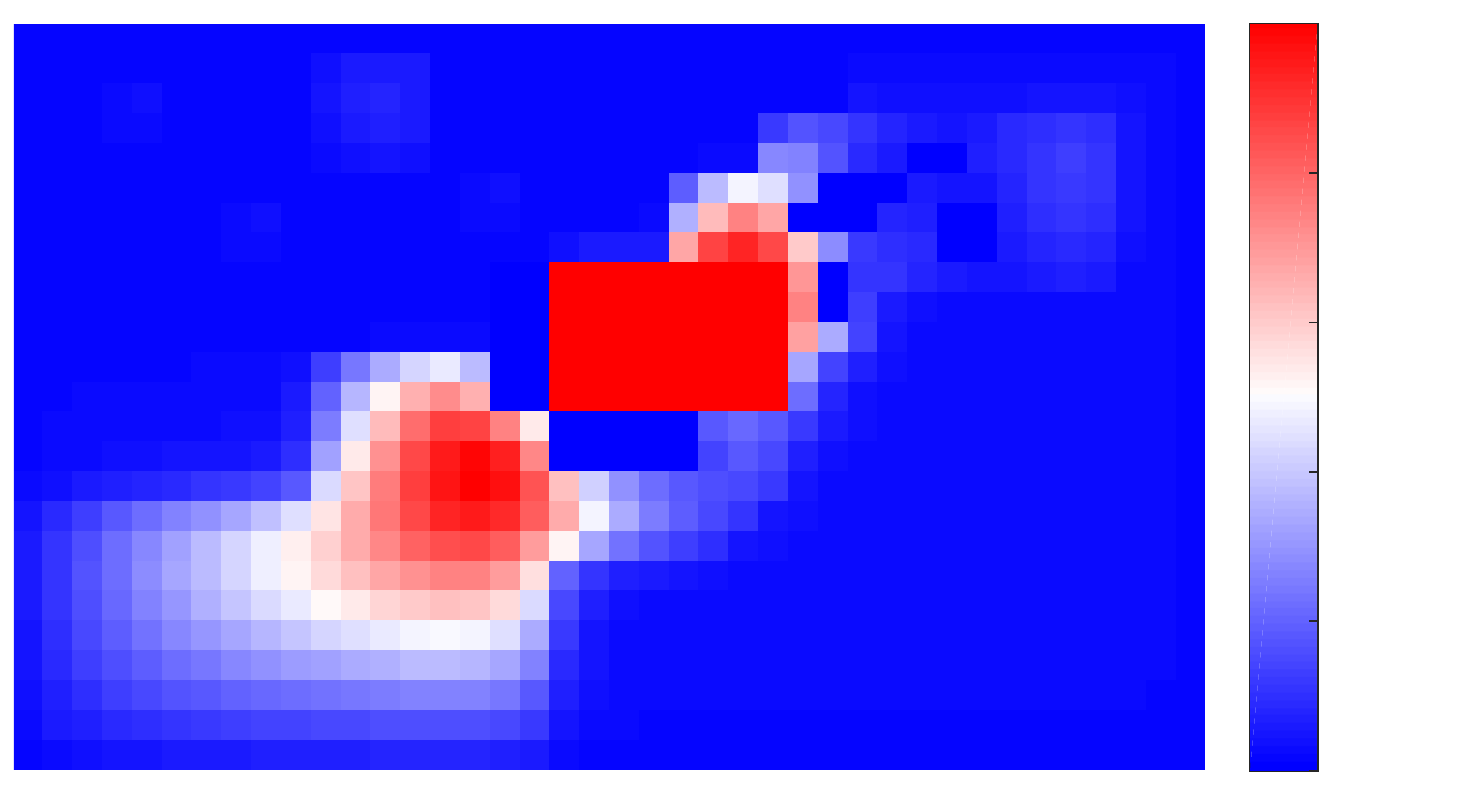
            \caption{Northeast}
            \label{fig:225}
    \end{subfigure}%
    \begin{subfigure}[b]{0.48\linewidth}
            \centering
            \def\svgwidth{0.99\linewidth}
              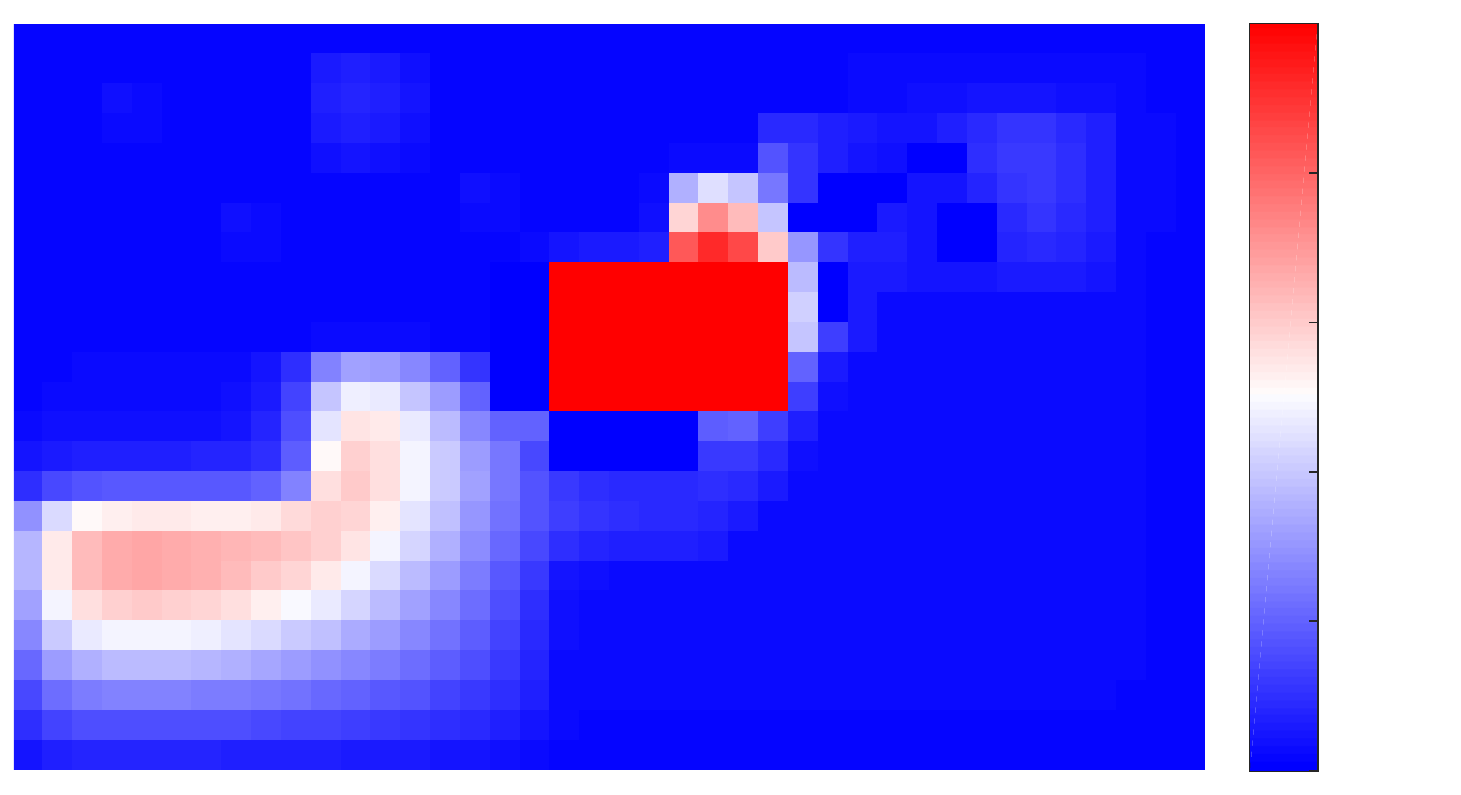
            \caption{Northwest}
            \label{fig:315}
    \end{subfigure}
    \caption{Normalized surveillance priority map for different wind directions for the landscape in Fig. \ref{fig:Ex2}. Wind direction is indicated as the direction the wind is coming from.}
\label{fig:SWind}
\end{figure}

\begin{figure}[!ht]
\centering
\begin{subfigure}[b]{0.48\linewidth}
        \includegraphics[width=1\linewidth]{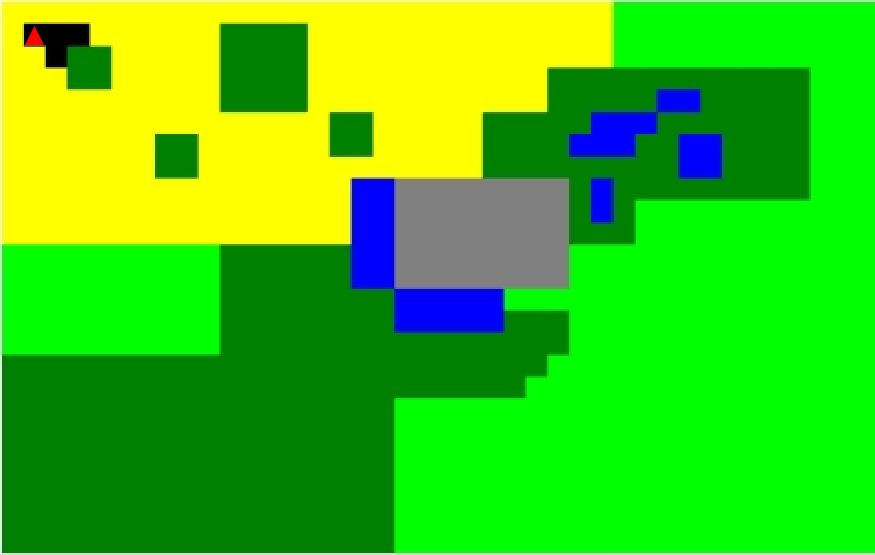}
      \caption{Low priority node}
      \label{fig:FireNW}
  \end{subfigure}
  ~ 
     \begin{subfigure}[b]{0.48\linewidth}
    \includegraphics[width=1\linewidth]{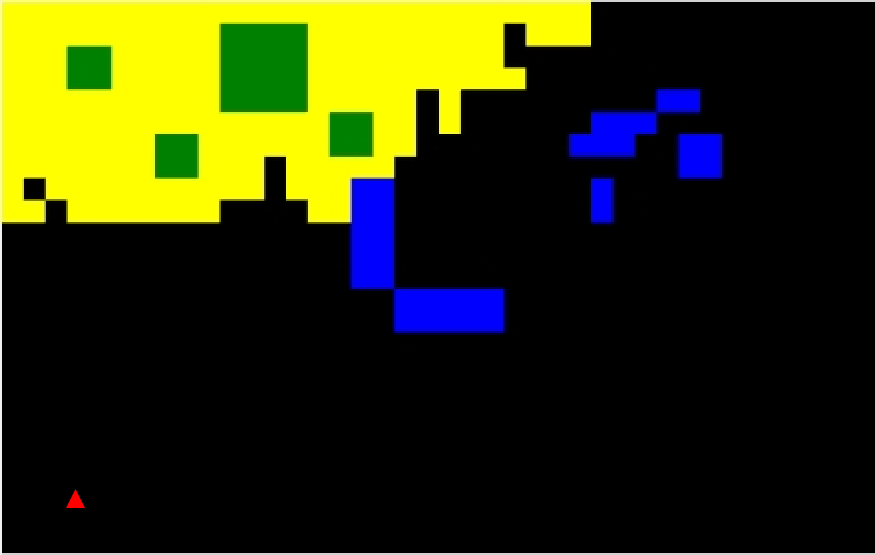}
      \caption{High priority node}
      \label{fig:FireSW}
  \end{subfigure}
      \caption{Final state $x(t_{f})$ of fire spread simulations of starting fires (red triangle) in different priority nodes for optimal wind conditions for fire spread to the city in the top and bottom left corner. Black indicates burned area. }
\label{fig:DifNodes}
\end{figure}

The spreading rate is furthermore adjusted for respectively a southwesterly, southeasterly, northeasterly and northwesterly wind with a speed of $V=4$ m/s. Taking a discount rate of  $r=5$, this results in Fig. \ref{fig:SWind}, where clearly the effects of wind direction on the node priority map can be seen. In general, the most important nodes are the nodes in the bottom left corner where the eucalyptic with grassland in between leads up to the city. Other possible hazards are the city itself due to its high cost and the smaller patches of eucalyptus closer to the city. However this area is smaller and restricted above by slow spreading desert and water. Therefore it poses less danger in case of fire igniting in the top part of the landscape. For better interpretation of the priority map, simulations were run starting fires in different priority nodes and the results can be seen in Fig. \ref{fig:DifNodes}.

\begin{figure}
\centering
    \begin{subfigure}[b]{0.48\linewidth}            
             \def\svgwidth{0.99\linewidth}
              \input{NWind45.pdf_tex}
            \caption{$c_{i}=0.01$}
            \label{fig:C001}
    \end{subfigure}%
    \begin{subfigure}[b]{0.48\linewidth}
            \centering
           \def\svgwidth{0.99\linewidth}
              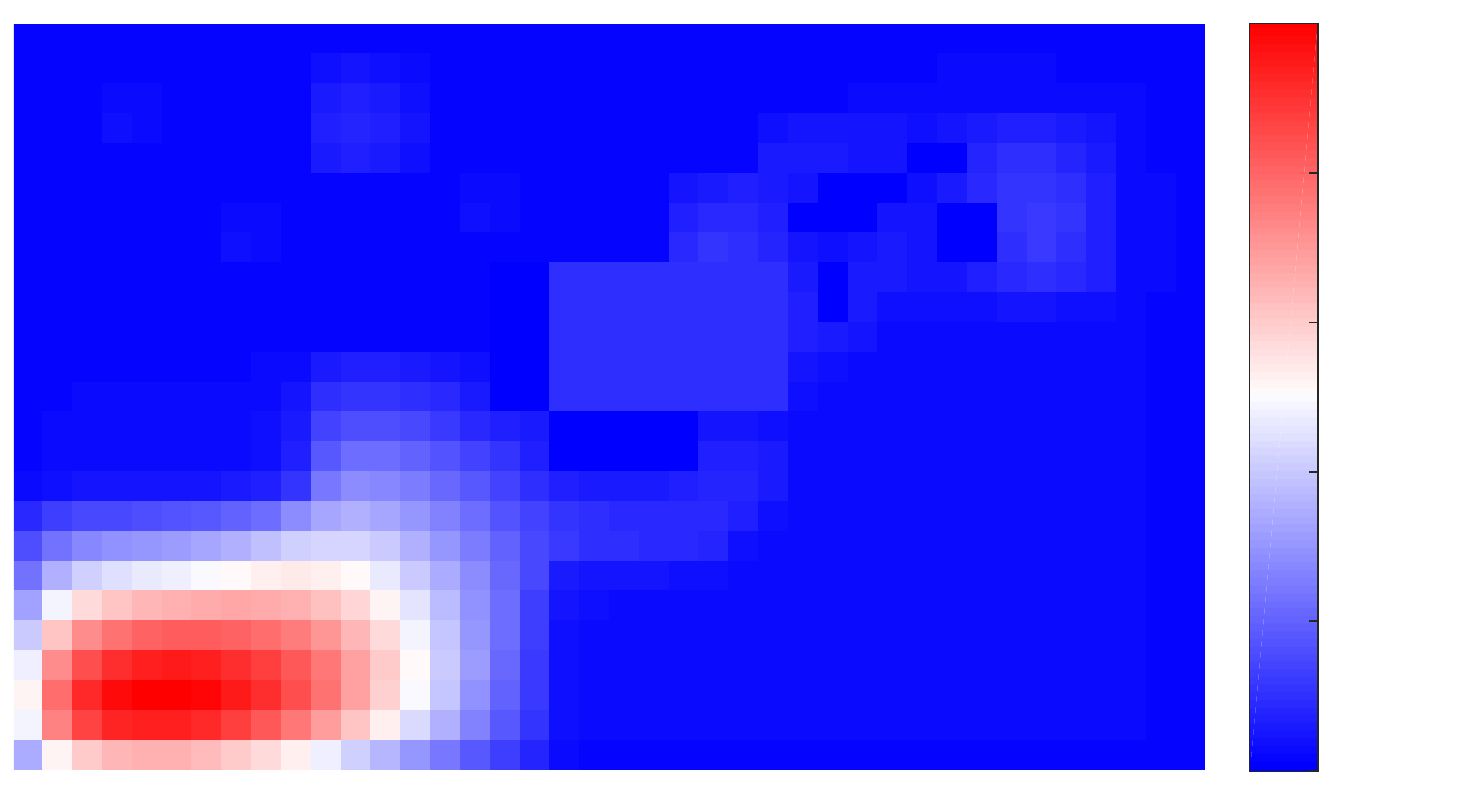
            \caption{$c_{i}=0.1$}
            \label{fig:C01}
    \end{subfigure}
    \caption{Effect of different cost $c_{i}$ of the landscape nodes on the priority map of Example 2 with southwesterly wind of $V=4$ m/s and $r=5$.}
\label{fig:CC}
\end{figure}

The cost that is taken for the landscape nodes also has an effect on the priority map as shown in Fig. \ref{fig:CC}. With a higher cost on the landscape nodes, they are prioritized over the city, because more landscape burning down will now result in a higher total cost.

%
%
%

\subsection{Discount Rate}
\begin{figure}
\centering
    \begin{subfigure}[b]{0.48\linewidth}            
             \def\svgwidth{0.99\linewidth}
              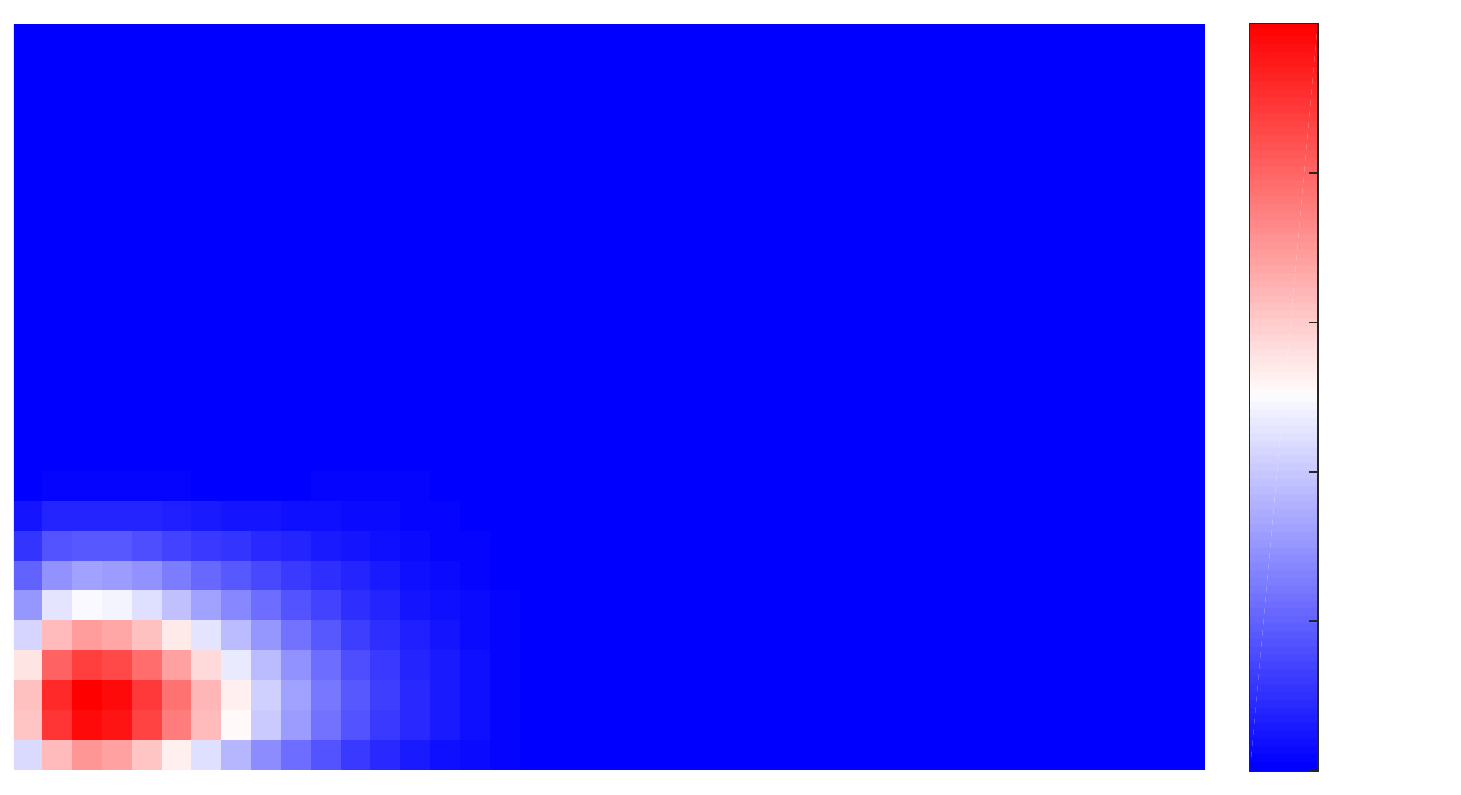
            \caption{$r=4.705$}
            \label{fig:R4705}
    \end{subfigure}%
    \begin{subfigure}[b]{0.48\linewidth}
            \centering
           \def\svgwidth{0.99\linewidth}
              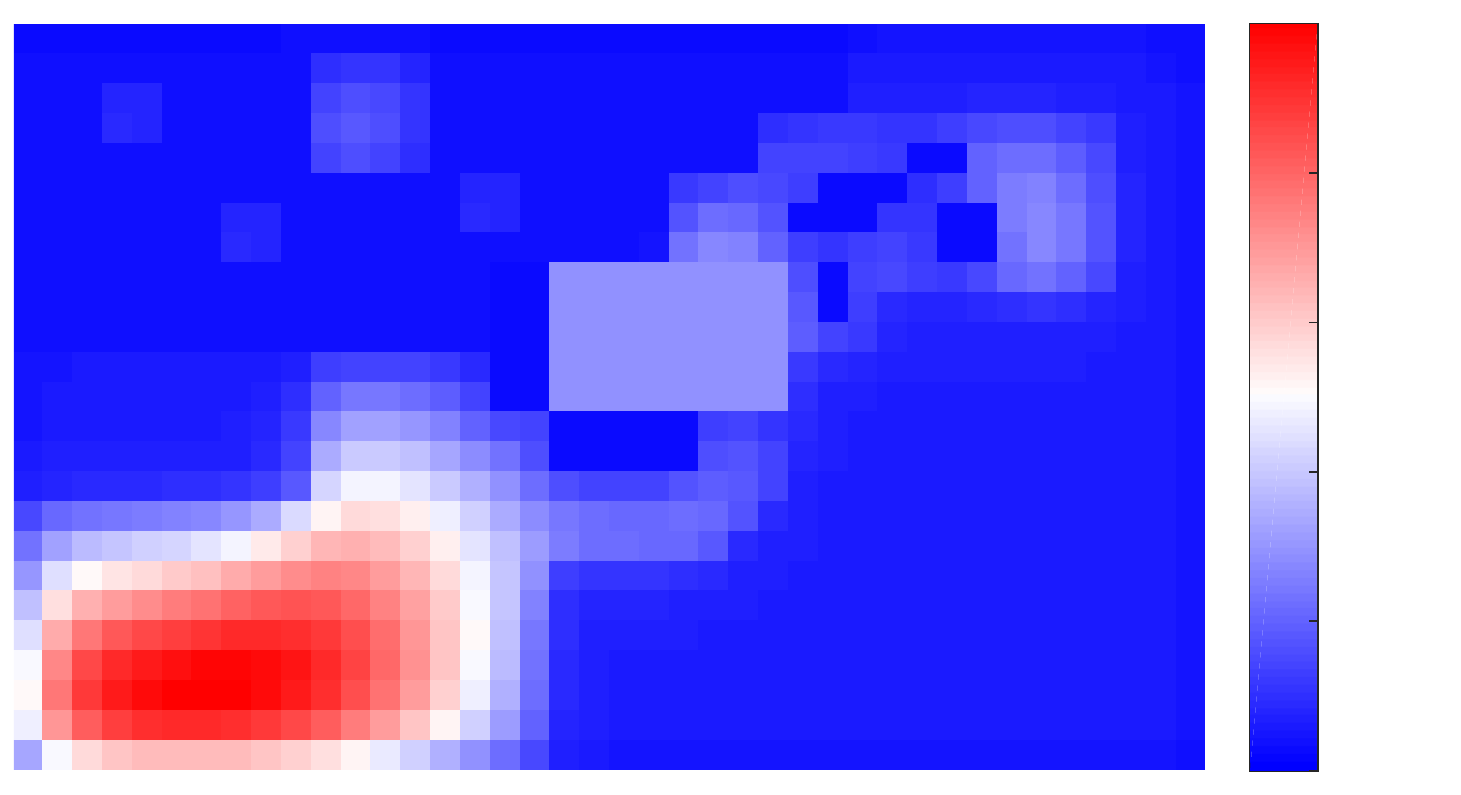
            \caption{$r=5.2$}
            \label{fig:R52}
    \end{subfigure}
    \begin{subfigure}[b]{0.48\linewidth}            
            \def\svgwidth{0.99\linewidth}
              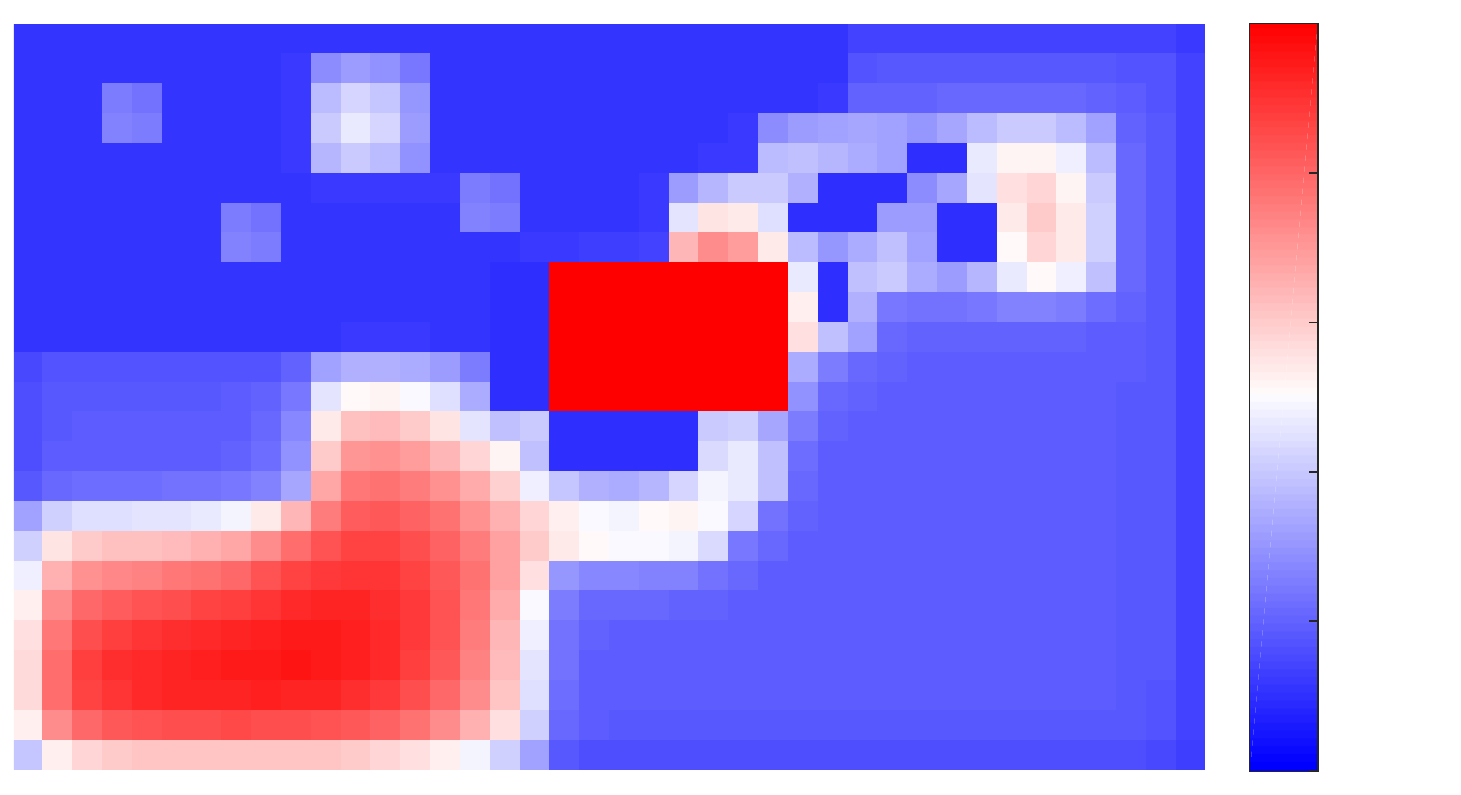
            \caption{$r=5.8$}
            \label{fig:R58}
    \end{subfigure}%
    \begin{subfigure}[b]{0.48\linewidth}
            \centering
            \def\svgwidth{0.99\linewidth}
              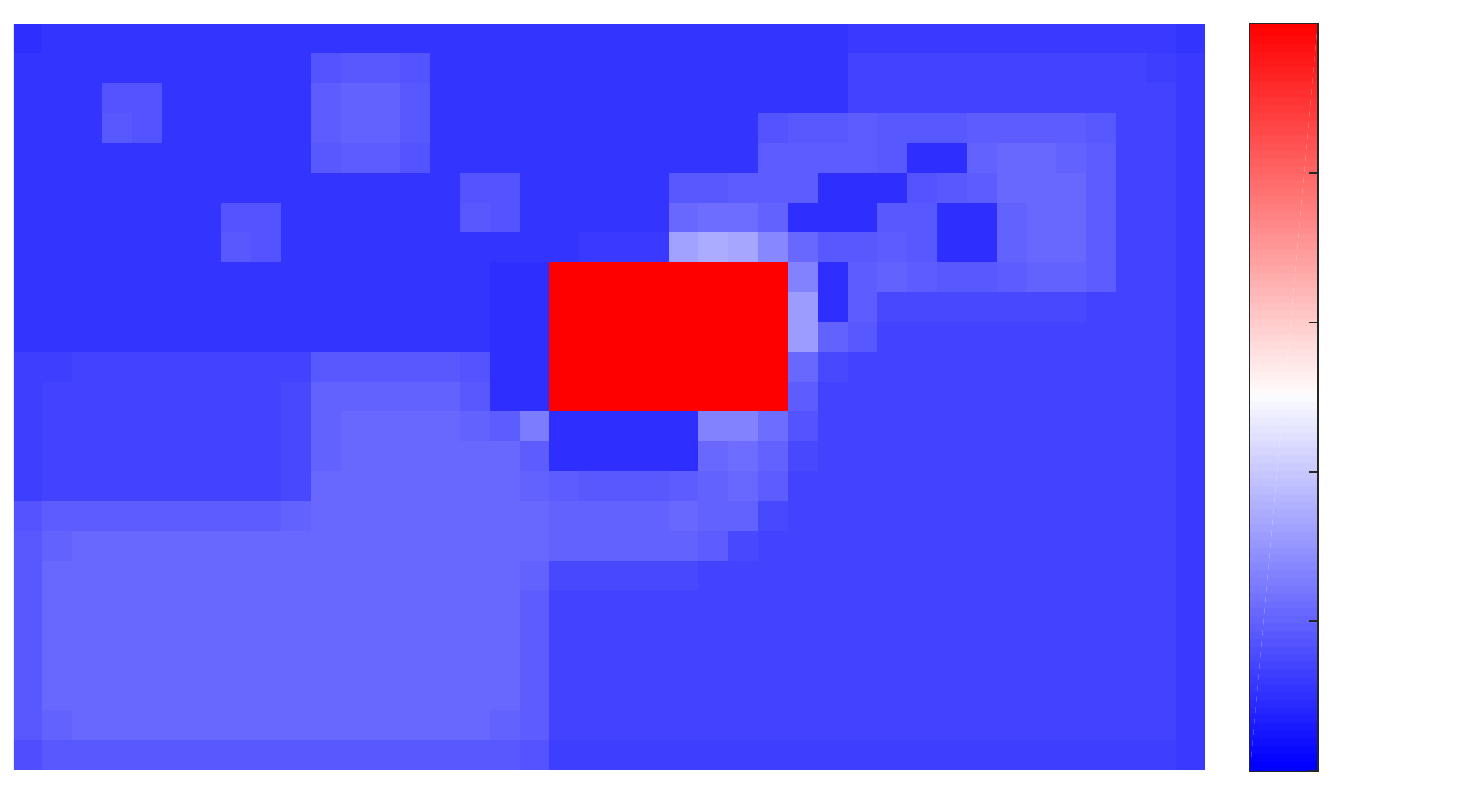
            \caption{$r=10$}
            \label{fig:R10}
    \end{subfigure}
    \caption{Effect of discount rate $r$ on the priority map of Example 2 with southwesterly wind of $V=4$ m/s and $c_{i}=0.1$ for non city nodes.}
\label{fig:DR}
\end{figure}

In section II we discussed the discount rate $r$ in (\ref{eq:cost2}). For Example 2 with $c_{i}=0.1$ for landscape nodes, $c_{i}=1$ for city nodes and a southwesterly wind of $V=4$ m/s, this effect is visualized in Fig. \ref{fig:DR}, where the minimal discount rate equals $r=4.705$ before $rI-A$ becomes singular. A higher discount rate prioritizes the near future and hence, prioritizes what is happening in the city, i.e. the highest cost area of the landscape, over what possible outbreaks in other parts of the graph could lead to. In the previously and future examples the discount rate is taken close to the minimal discount rate, i.e. the positive real part of the largest eigenvalue of A, while still attributing some importance to the city, to sketch a more interesting and complete image of the situation.


\section{INTERVENTION RESULTS}
Let us consider the same examples as in Section III, however now the task is not to prioritize nodes, but the optimal control problem of lowering the total system cost by changing the systems dynamics in the most effective way as explained in Section \ref{subsec:IP}. Compared to Section III, the results do not directly represent the priority of the nodes, but give a priority on which links control resources should be applied to for the minimal cost of an outbreak in the system. The obtained map will therefore, henceforth be called a cost-to-go map to distinguish the difference with the previously obtained priority map.

\subsection{Example 1}
\begin{figure}
\centering
    \begin{subfigure}[b]{0.48\linewidth}            
           \def\svgwidth{0.99\linewidth}
              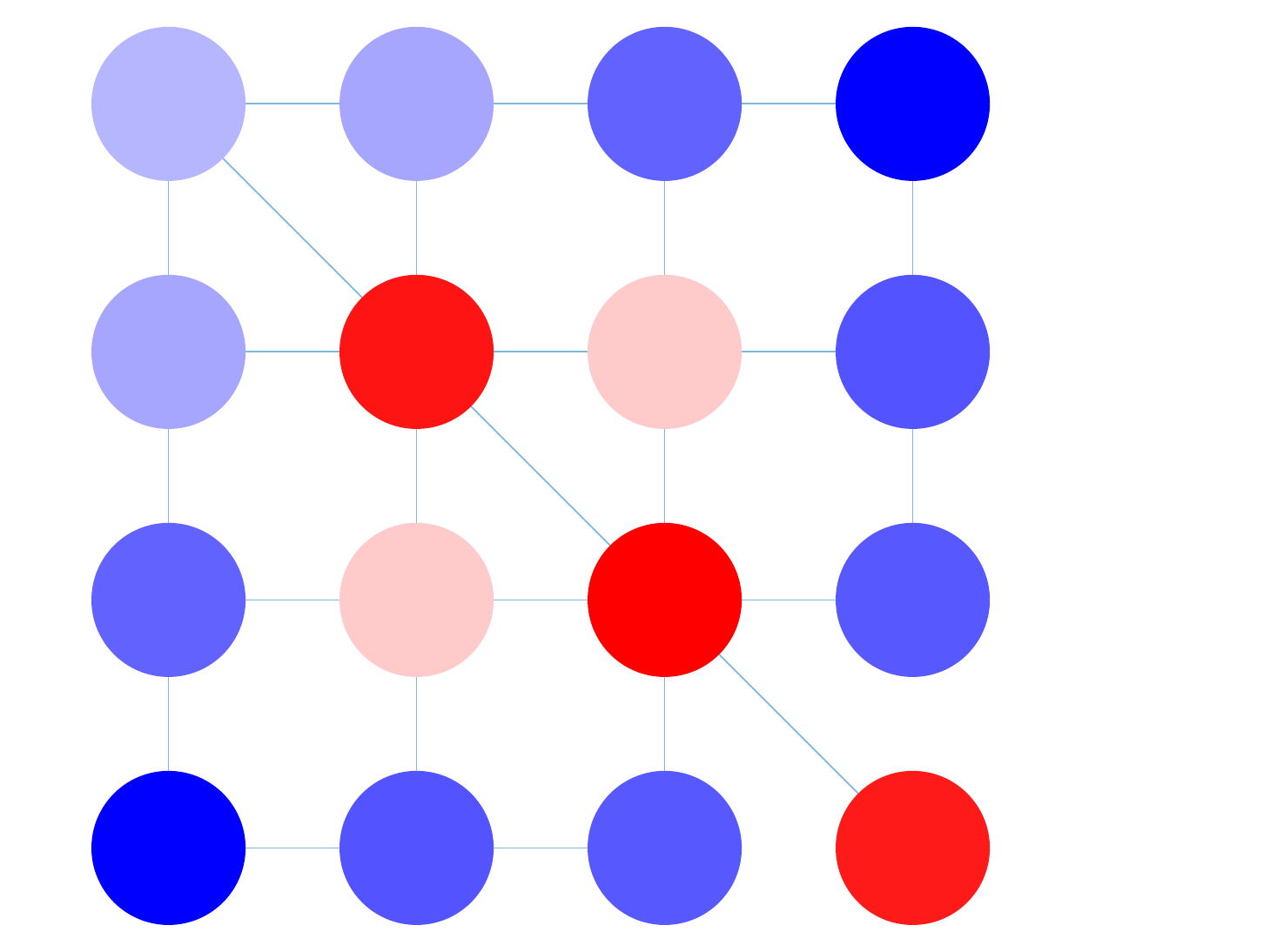
            \caption{$\Gamma=0$}
            \label{fig:IG0}
    \end{subfigure}%
    \begin{subfigure}[b]{0.48\linewidth}
            \centering
             \def\svgwidth{0.99\linewidth}
              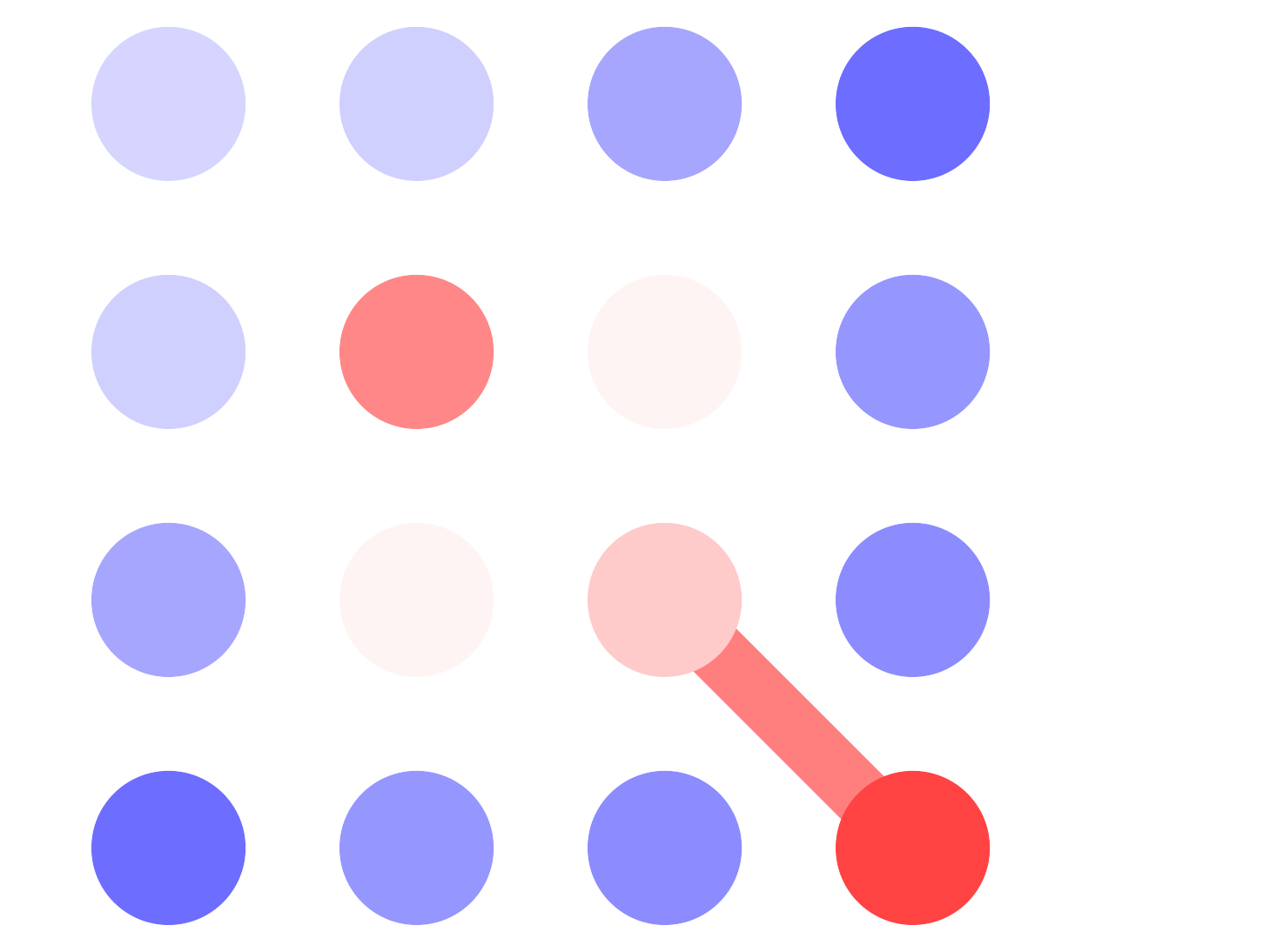
            \caption{$\Gamma=0.5$}
            \label{fig:IG05}
    \end{subfigure}
    \begin{subfigure}[b]{0.48\linewidth}
            \centering
            \def\svgwidth{0.99\linewidth}
              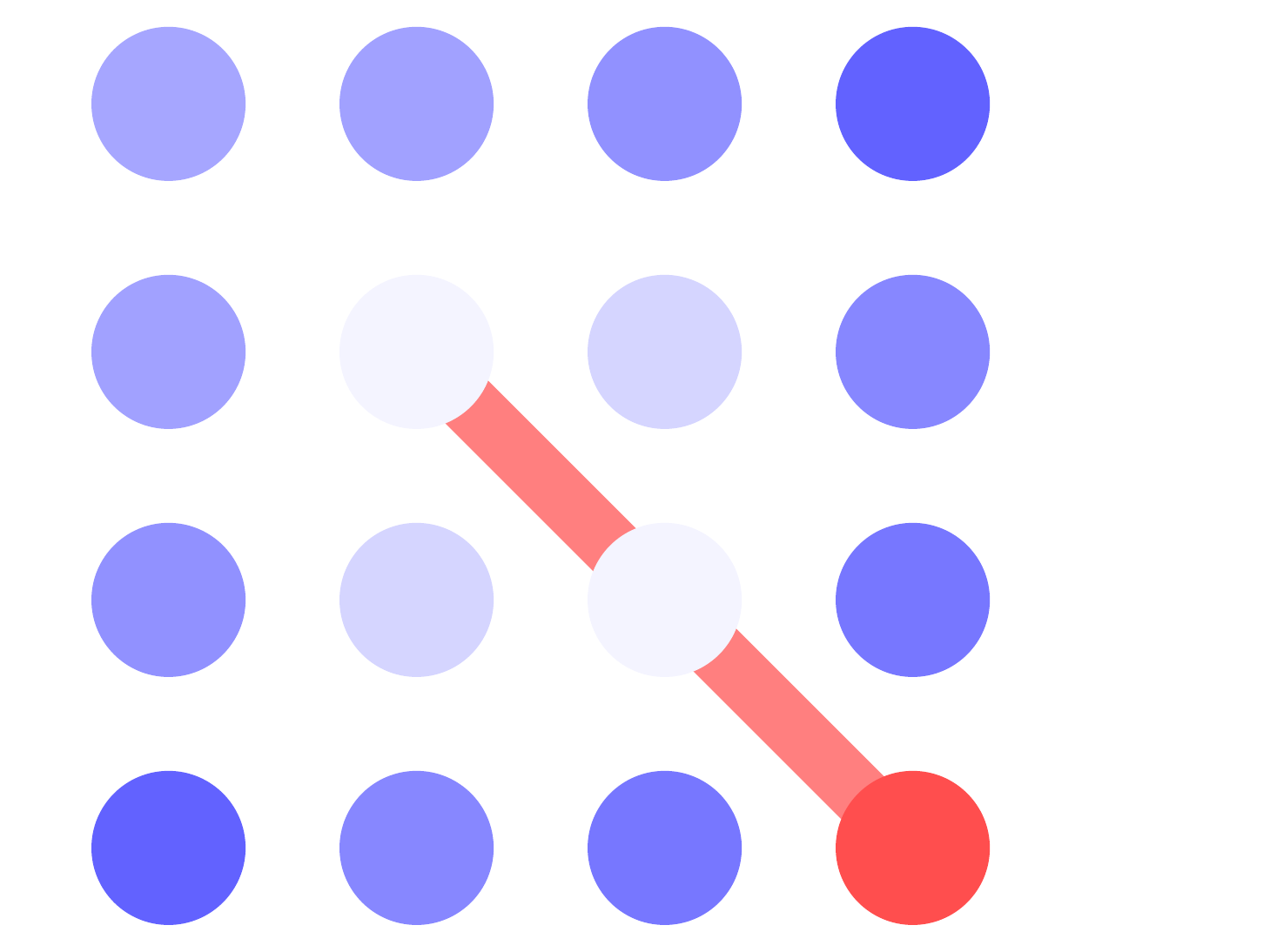
            \caption{$\Gamma=1$}
            \label{fig:IG1}
    \end{subfigure}
    \begin{subfigure}[b]{0.48\linewidth}
            \centering
            \def\svgwidth{0.99\linewidth}
              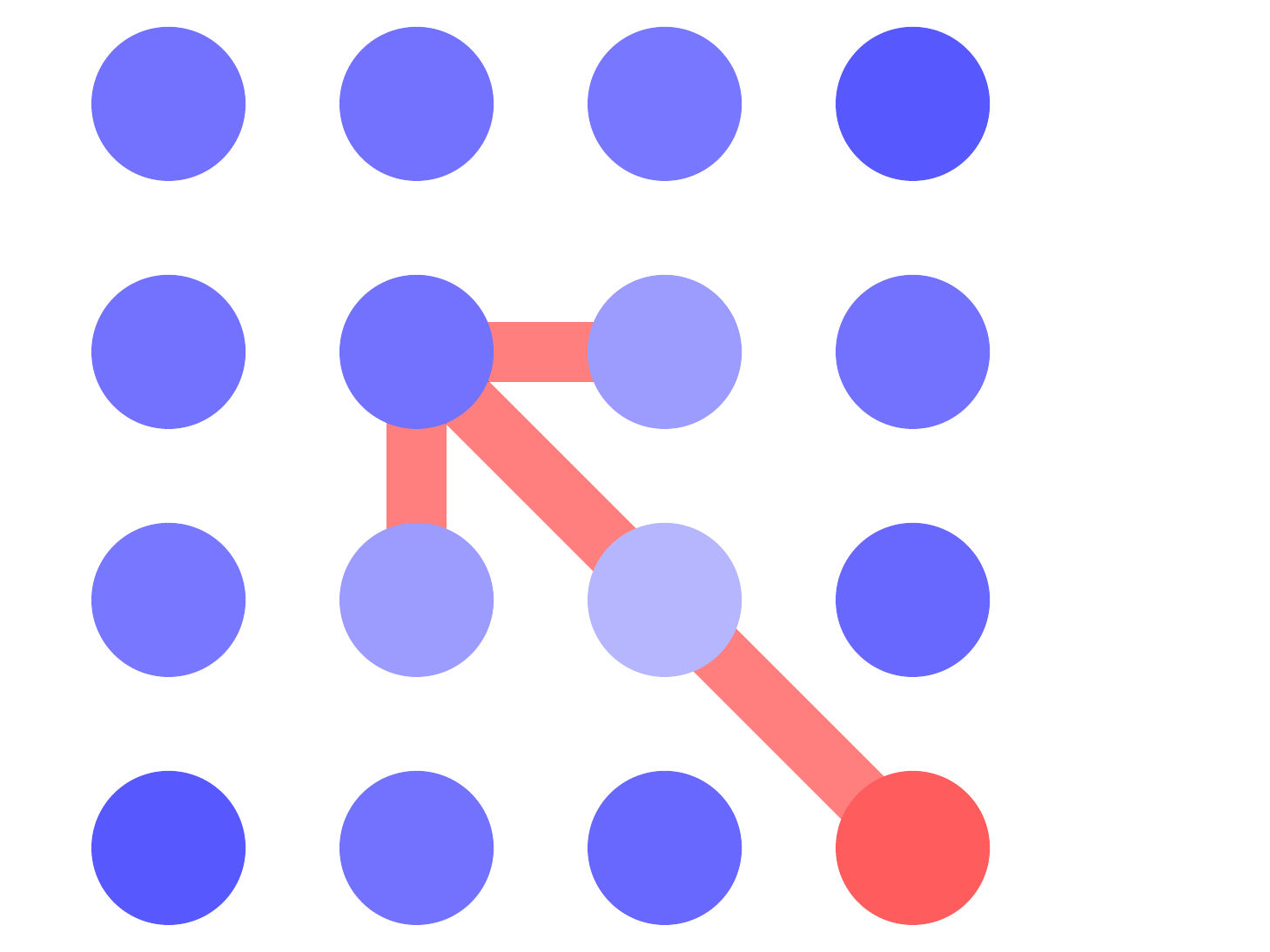
            \caption{$\Gamma=2$}
            \label{fig:IG2}
    \end{subfigure}
    \caption{Cost-to-go map for the intervention problem belonging to the graph displayed in Fig. \ref{fig:Graph1} for different values of $\Gamma$. The color and width of the links represent the amount of control resources to be applied, whereas the node color represents the cost-to-go. }
\label{fig:IEx1}
\end{figure}

We consider the graph with $n=16$ nodes from Example 1, displayed in Fig. \ref{fig:Graph1}, with recovery rate $\delta=0.2$ and equal spreading rate of $\beta=0.5$ for connected nodes. In Fig. \ref{fig:IEx1} the obtained control resource allocation is demonstrated by the color and width of the links. The node colors represent the updated priority or cost-to-go of the nodes, with the same color-scale as the case without intervention, i.e. $\Gamma=0$. 

It is clear that increasing $\Gamma$ reduces the cost-to-go of the critical nodes. Furthermore, the allocation of resources is very sparse, which is well-suited to waterbombing or controlled-burning interventions. It can be seen that for $\Gamma=0.5$, applying all control resources to the link connecting diagonal node $11$ with the city node $16$ reduces the overall cost-to-go of the nodes significantly. This can be interpreted as the overall effect or outcome of one or more of these particular nodes being set on fire is reduced, i.e. the total cost will be less. For more resources available, more links are targeted but still in a sparse manner. 

The effect of the discount rate $r$ on the distribution of the control resources is similar to the effect on the priority maps, where for long-term reduction in cost it is best to intervene at a central point, but for short-term focused optimisation it is best to intervene close to the city.

\begin{figure}
\centering
    \begin{subfigure}[b]{0.48\linewidth}            
           \def\svgwidth{0.99\linewidth}
              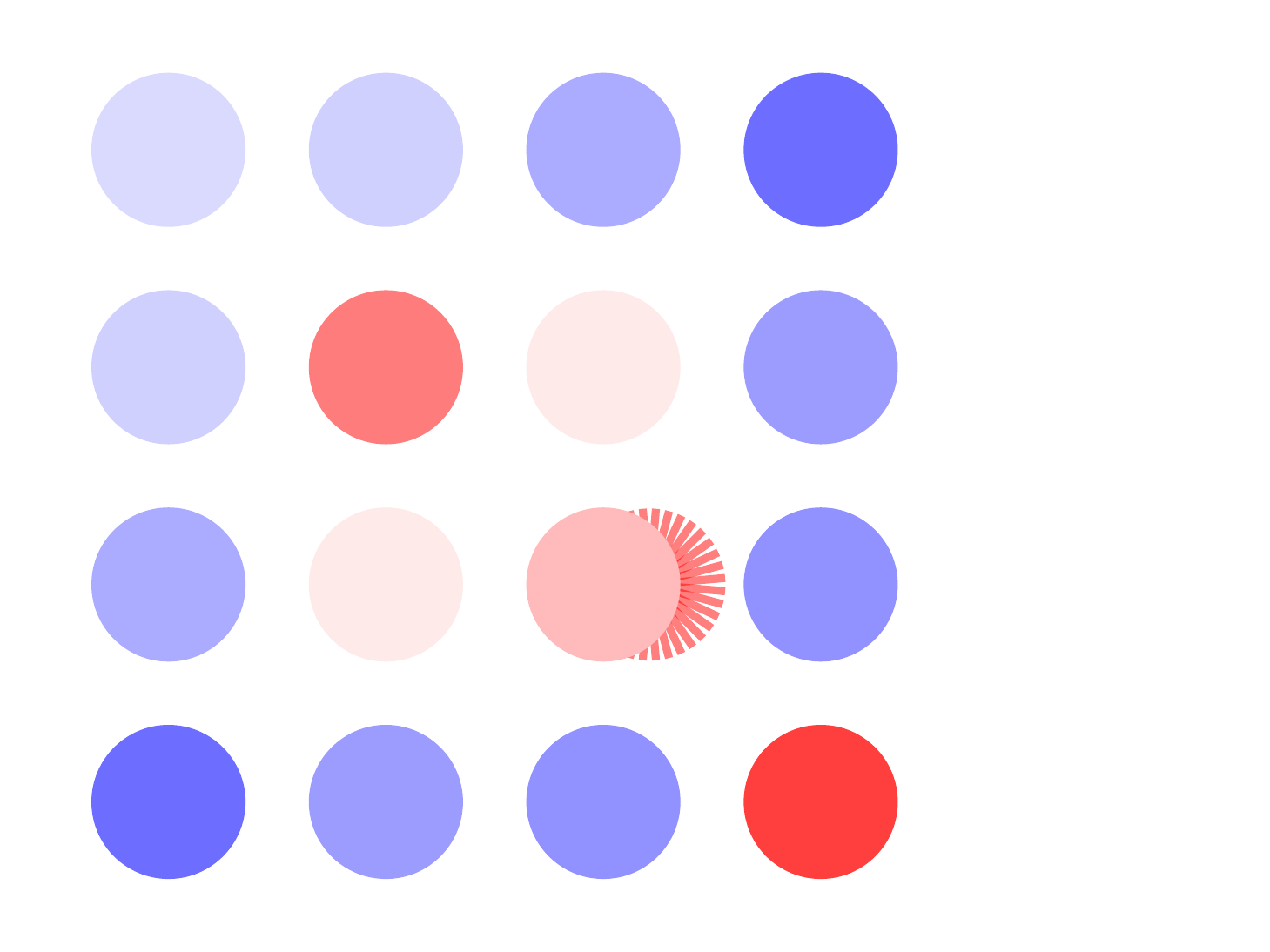
            \caption{$\Gamma=0.5$}
            \label{fig:IDG05}
    \end{subfigure}%
    \begin{subfigure}[b]{0.48\linewidth}
            \centering
             \def\svgwidth{0.99\linewidth}
              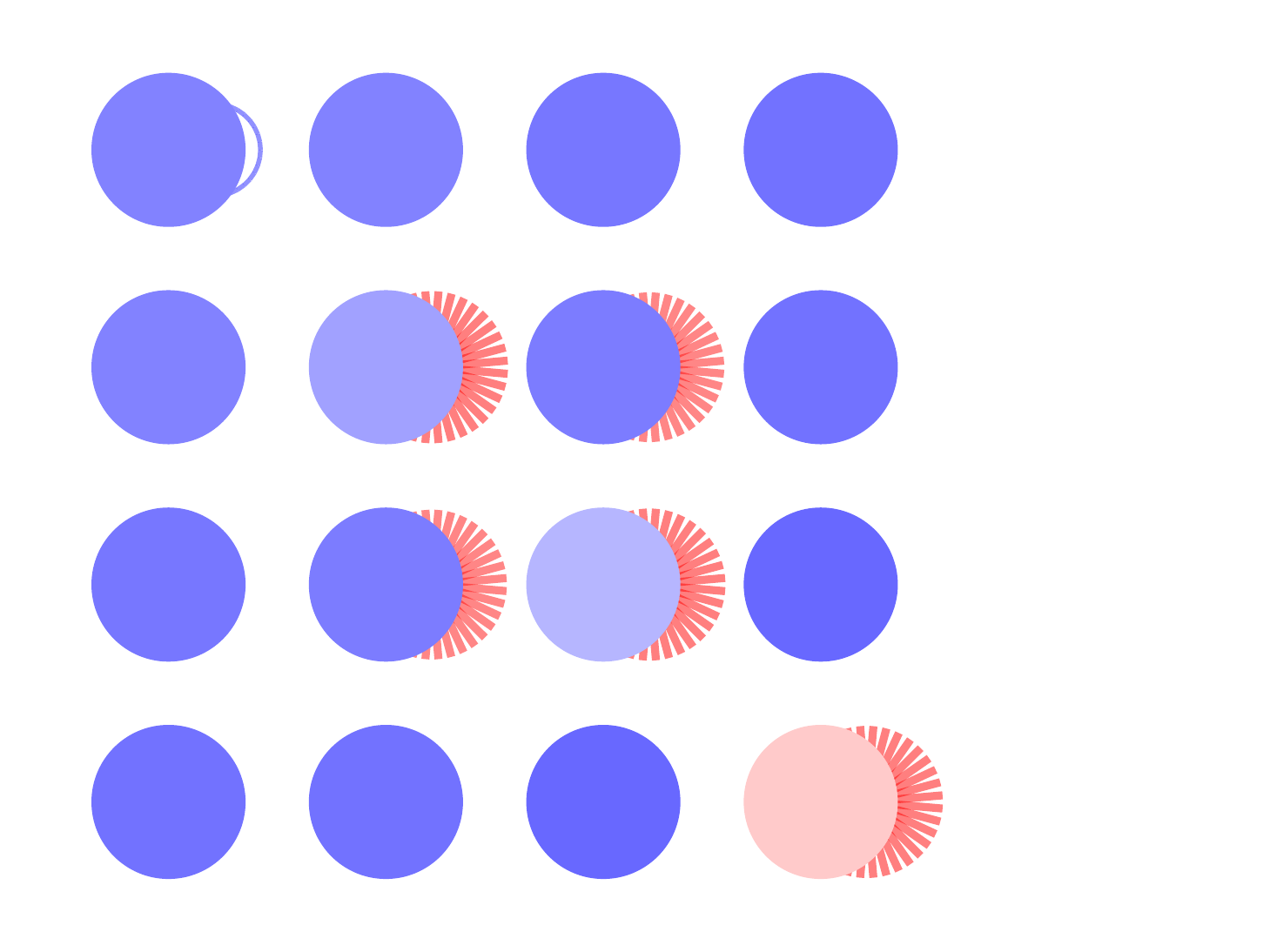
            \caption{$\Gamma=4$}
            \label{fig:IDG4}
    \end{subfigure}
    \caption{Cost-to-go map for the intervention problem belonging to the graph in Fig. \ref{fig:Graph1} if control resources, indicated as red crescents, are applied to nodes, i.e. $k_{ii}$, for $r=2$. }
\label{fig:Diag}
\end{figure}

Now, taking the diagonal entries of K as nonzero and hence applying control resources to the nodes rather than edges, results in Fig. \ref{fig:Diag}. The allocation of resources is similar to Fig. \ref{fig:IEx1}.

\subsection{Example 2}
\begin{figure}[!ht]
  \centering
    \def\svgwidth{\linewidth}
              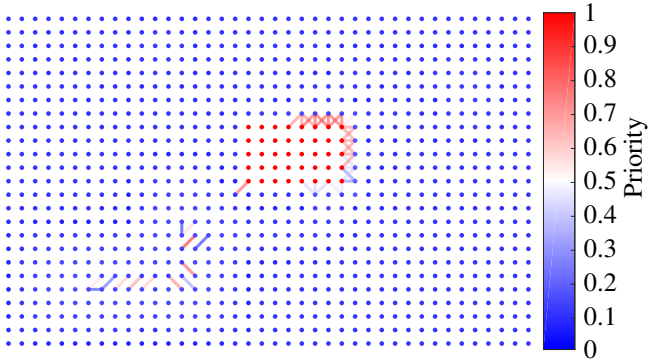
    \caption{Cost-to-go map with control resources on the links for landscape Example 2, displayed in Fig. \ref{fig:Ex2}, for $\Gamma=40$.}
    \label{fig:IEx2}
\end{figure}

To show the intervention problem is also scalable for larger examples, the intervention map for Example 2, with spreading dynamics as described in Section III-B, no wind present, $c_{i}=0.01$, $\Gamma=40$ and $r=11.5$, is displayed in Fig. \ref{fig:IEx2}.  Again, it can be seen that the map of edges chosen for intervention is very sparse.  

The intervention map behaves similar to Example 1 regarding the amount of resources, cost and discount rate. I.e. depending on the amount of resources available more resources are applied to the links connecting the city and to the eucalyptus forest to the southeast. With less resources available this forest is of less priority due to it being connected to the city by only a single link. Finally, taking a larger cost $c_{i}$ on the landscape nodes or a lower discount rate $r$ would result in a higher cost-to-go on the landscape and therefore more resources would be dedicated to the landscape at the expense of the links connecting the city. 

\begin{figure}[!ht]
\centering
\begin{subfigure}[b]{0.48\linewidth}
        \includegraphics[width=1\linewidth]{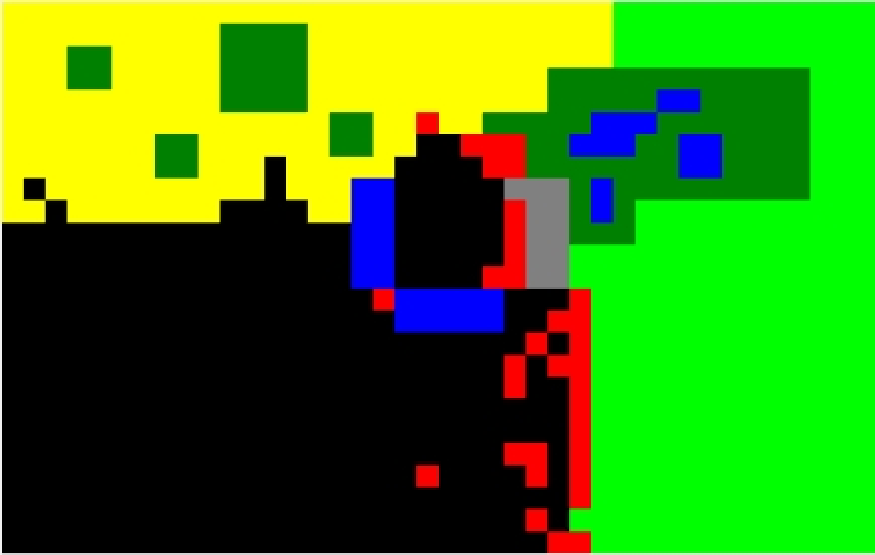}
      \caption{No Intervention, $x(0.5t_{f})$}
      \label{fig:NIhalf}
  \end{subfigure}
  ~ 
     \begin{subfigure}[b]{0.48\linewidth}
    \includegraphics[width=1\linewidth]{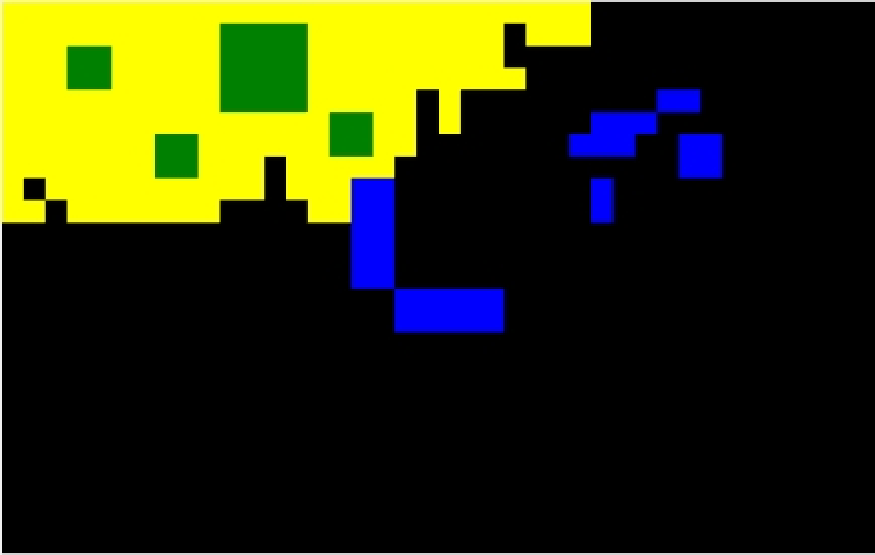}
      \caption{No Intervention, $x(t_{f})$}
      \label{fig:NIcomplete}
  \end{subfigure}
     \begin{subfigure}[b]{0.48\linewidth}
        \includegraphics[width=1\linewidth]{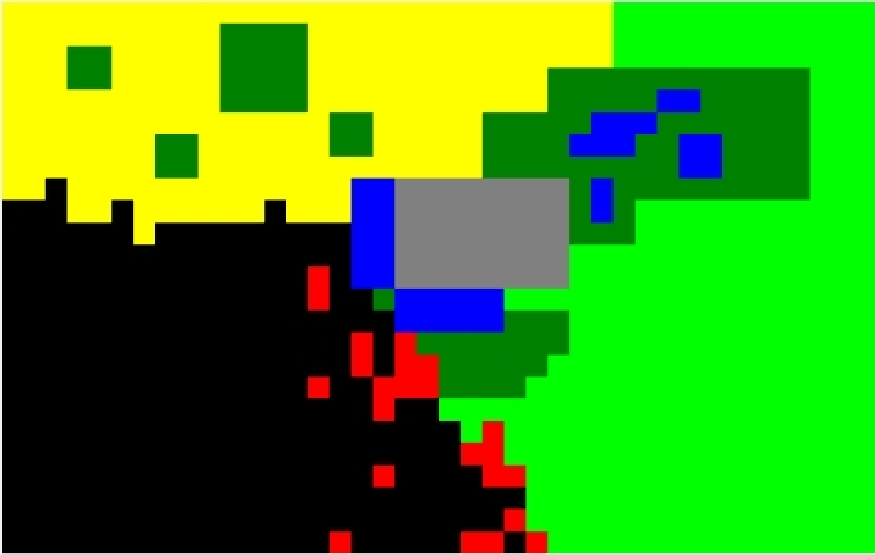}
      \caption{Intervention, $x(0.5t_{f})$}
      \label{fig:Ihalf}
  \end{subfigure}
    ~ 
     \begin{subfigure}[b]{0.48\linewidth}
        \includegraphics[width=1\linewidth]{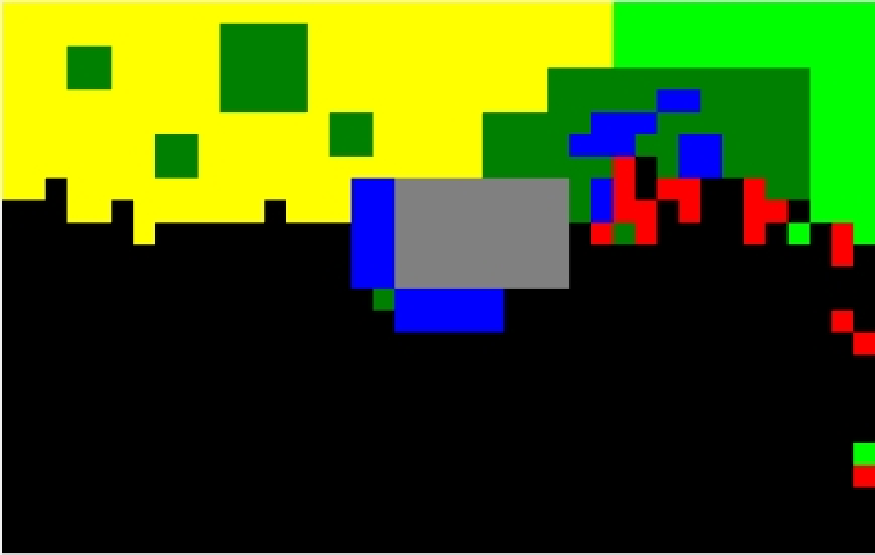}
      \caption{Intervention, $x(t_{f})$}
      \label{fig:Icomplete}
  \end{subfigure}
      \caption{Fire spread simulation, cf. Fig \ref{fig:FireSW}, with and without the intervention indicated in Fig. \ref{fig:IEx2}.} 
\label{fig:InterventionRun}
\end{figure}

The results of applying the interventions suggested in Fig. \ref{fig:IEx2} are demonstrated in Fig. \ref{fig:InterventionRun}. In case the control resources are applied, the city does not burn down and the fire spreads significantly more slowly. 

\subsection{Computation and Convergence} 
The surveillance problem can be directly solved by using (\ref{eq:MM}) and solves within 0.04 seconds for all given examples on a standard desktop computer. 

For the intervention problem on the other hand, to solve the linear program and objective as stated in Section \ref{subsec:IP}, multiple iterations are required before $P_{0}=P$. For both Example 1 and Example 2 this happens quickly within 5 to 10 iterations. The problem is solved with MOSEK in Matlab and for Example 1, with $n=16$ nodes, takes on average 0.1 second per iteration, whereas for Example 2, $n=1000$ nodes, each iteration takes up to 35 seconds. 
%

\subsection{Application to UAV Path Planning}
\begin{figure}
  \centering
    \def\svgwidth{0.9\linewidth}
              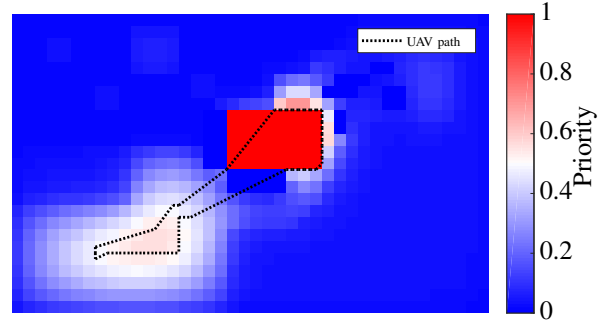
    \caption{Possible UAV path for Fig. \ref{fig:IEx2} targeting all critical links solving the traveling salesman problem, visualized on the $\Gamma=0$ priority map.}
    \label{fig:TSP}
\end{figure}

The main motivation for the developed methods is to obtain priority maps for UAV path planning purposes. Considering different constrained budgets (e.g. flight time, endurance, speed, sensor range) UAVs will have to make a trade-off where to fly if they are used for surveillance or intervention purposes. For the intervention problem, multiple targets have been identified that all have to be visited or hit. Therefore solving the traveling salesman problem is suitable and in Fig. \ref{fig:TSP} a possible UAV round trip is generated using binary integer programming for the intervention cost-to-go map for Example 2.  
For surveillance problems with fuel-limited and turn-rate-limited UAVs it will also be interesting to consider path-planning algorithms such as \cite{savla2008traveling, penicka2017dubins}, which we leave as future work.

\section{CONCLUSIONS}

In this paper we presented a method to obtain priority maps for spreading processes which can be used as an input for UAV path planning problems. Two different problems were considered: the surveillance problem and the intervention problem. Future work will include more testing with more realistic propagation models and sensing assumptions, and further integration with robot motion planning algorithms.






%
%
%


%
%
%
\bibliographystyle{IEEEtran}
\bibliography{IEEEabrv,ACFR02082018}

\begin{thebibliography}{10}
\providecommand{\url}[1]{#1}
\csname url@samestyle\endcsname
\providecommand{\newblock}{\relax}
\providecommand{\bibinfo}[2]{#2}
\providecommand{\BIBentrySTDinterwordspacing}{\spaceskip=0pt\relax}
\providecommand{\BIBentryALTinterwordstretchfactor}{4}
\providecommand{\BIBentryALTinterwordspacing}{\spaceskip=\fontdimen2\font plus
\BIBentryALTinterwordstretchfactor\fontdimen3\font minus
  \fontdimen4\font\relax}
\providecommand{\BIBforeignlanguage}[2]{{%
\expandafter\ifx\csname l@#1\endcsname\relax
\typeout{** WARNING: IEEEtran.bst: No hyphenation pattern has been}%
\typeout{** loaded for the language `#1'. Using the pattern for}%
\typeout{** the default language instead.}%
\else
\language=\csname l@#1\endcsname
\fi
#2}}
\providecommand{\BIBdecl}{\relax}
\BIBdecl

\bibitem{Messinger2016}
M.~Messinger and M.~Silman, ``{Unmanned aerial vehicles for the assessment and
  monitoring of environmental contamination: An example from coal ash
  spills},'' \emph{Environmental Pollution}, vol. 218, pp. 889--894, 2016.

\bibitem{Mcclean2010}
W.~T.~L. Teacy, J.~Nie, S.~Mcclean, G.~Parr, S.~Hailes, and S.~Julier,
  ``{Collaborative Sensing by Unmanned Aerial Vehicles},'' in \emph{the 3rd
  International Workshop on Agent Technology for Sensor Networks}, 2009.

\bibitem{AmericanRedCross2015}
{American Red Cross}, ``{Drones for Disaster Response and Relief Operations},''
  Tech. Rep. April, 2015.

\bibitem{Howden2008}
D.~Howden and T.~Hendtlass, ``{Collective Intelligence and Bush Fire
  Spotting},'' in \emph{GECCO '08: Proceedings of the 10th annual conference on
  Genetic and evolutionary computation}, 2008, pp. 41--48.

\bibitem{Nigam2014}
N.~Nigam, ``{The Multiple Unmanned Air Vehicle Persistent Surveillance Problem:
  A Review},'' \emph{Machines}, vol.~2, no.~1, pp. 13--72, 2014.

\bibitem{Karafyllidis1997a}
I.~Karafyllidis and A.~Thanailakis, ``{A model for predicting forest fire
  spreading using cellular automata},'' \emph{Ecological Modelling}, vol.~99,
  no.~1, pp. 87--97, 1997.

\bibitem{Johnston2006}
P.~Johnston, G.~Milne, and J.~Kelso, ``{A heat transfer simulation model for
  wildfire spread},'' \emph{Forest Ecology and Management}, vol. 234, p. S78,
  2006.

\bibitem{yu2016correlated}
J.~Yu, M.~Schwager, and D.~Rus, ``Correlated orienteering problem and its
  application to persistent monitoring tasks,'' \emph{IEEE Transactions on
  Robotics}, vol.~32, no.~5, pp. 1106--1118, 2016.

\bibitem{smith2012persistent}
S.~L. Smith, M.~Schwager, and D.~Rus, ``Persistent robotic tasks: Monitoring
  and sweeping in changing environments,'' \emph{IEEE Transactions on
  Robotics}, vol.~28, no.~2, pp. 410--426, 2012.

\bibitem{penicka2017dubins}
R.~Penicka, J.~Faigl, P.~V{\'a}na, and M.~Saska, ``Dubins orienteering
  problem.'' \emph{IEEE Robotics and Automation Letters}, vol.~2, no.~2, pp.
  1210--1217, 2017.

\bibitem{Nowzari2015}
C.~Nowzari, V.~M. Preciado, and G.~J. Pappas, ``{Analysis and Control of
  Epidemics: A Survey of Spreading Processes on Complex Networks},'' \emph{IEEE
  Control Systems}, vol.~36, no.~1, pp. 26--46, 2016.

\bibitem{Rothermel1972}
R.~C. Rothermel, ``{A mathematical model for predicting fire spread in wildland
  fuels},'' USDA, Oregon, Tech. Rep., 1972.

\bibitem{Rothermel1983New}
------, ``How to predict the spread and intensity of forest and range fires,''
  USDA, Tech. Rep., 1983.

\bibitem{Sullivan2009}
A.~L. Sullivan, ``{A review of wildland fire spread modelling, 1990-present 3:
  Mathematical analogues and simulation models},'' \emph{International Journal
  of Wildland Fire}, vol.~18, pp. 387--403, 2007.

\bibitem{Alexandridis2008a}
A.~Alexandridis, D.~Vakalis, C.~I. Siettos, and G.~V. Bafas, ``{A cellular
  automata model for forest fire spread prediction: The case of the wildfire
  that swept through Spetses Island in 1990},'' \emph{Applied Mathematics and
  Computation}, vol. 204, no.~1, pp. 191--201, 2008.

\bibitem{Kelso2015}
J.~K. Kelso, D.~Mellor, M.~E. Murphy, and G.~J. Milne, ``{Techniques for
  evaluating wildfire simulators via the simulation of historical fires using
  the Australis simulator},'' \emph{International Journal of Wildland Fire},
  vol.~24, no.~6, pp. 784--797, 2015.

\bibitem{kermark1927contributions}
M.~Kermark and A.~Mckendrick, ``Contributions to the mathematical theory of
  epidemics. part i,'' \emph{Proc. r. soc. a}, vol. 115, no.~5, pp. 700--721,
  1927.

\bibitem{bailey1975mathematical}
N.~T. Bailey \emph{et~al.}, \emph{The mathematical theory of infectious
  diseases and its applications}.\hskip 1em plus 0.5em minus 0.4em\relax
  Charles Griffin \& Company Ltd, 1975.

\bibitem{Ahn2013}
H.~J. Ahn and B.~Hassibi, ``Global dynamics of epidemic spread over complex
  networks,'' in \emph{52nd IEEE Conference on Decision and Control}, Dec 2013,
  pp. 4579--4585.

\bibitem{VanMieghem:2009}
P.~Van~Mieghem, J.~Omic, and R.~Kooij, ``Virus spread in networks,''
  \emph{IEEE/ACM Trans. Netw.}, vol.~17, no.~1, pp. 1--14, Feb. 2009.

\bibitem{Nowzari2016}
C.~Nowzari, V.~M. Preciado, and G.~J. Pappas, ``{Analysis and Control of
  Epidemics: A Survey of Spreading Processes on Complex Networks},'' \emph{IEEE
  Control Systems}, vol.~36, no.~1, pp. 26--46, 2016.

\bibitem{Madar2004}
N.~Madar, T.~Kalisky, R.~Cohen, D.~Ben-Avraham, and S.~Havlin, ``{Immunization
  and epidemic dynamics in complex networks},'' \emph{The European Physical
  Journal B - Condensed Matter and Complex Systems}, vol.~38, no.~2, pp.
  269--276, 2004.

\bibitem{youssef2011individual}
M.~Youssef and C.~Scoglio, ``An individual-based approach to sir epidemics in
  contact networks,'' \emph{Journal of theoretical biology}, vol. 283, no.~1,
  pp. 136--144, 2011.

\bibitem{Liu2016}
J.~G. Liu, J.~H. Lin, Q.~Guo, and T.~Zhou, ``{Locating influential nodes via
  dynamics-sensitive centrality},'' \emph{Scientific Reports}, vol.~6, 2016.

\bibitem{van2011decreasing}
P.~Van~Mieghem, D.~Stevanovi{\'c}, F.~Kuipers, C.~Li, R.~Van De~Bovenkamp,
  D.~Liu, and H.~Wang, ``Decreasing the spectral radius of a graph by link
  removals,'' \emph{Physical Review E}, vol.~84, no.~1, p. 016101, 2011.

\bibitem{Preciado}
V.~M. Preciado, M.~Zargham, C.~Enyioha, A.~Jadbabaie, and G.~Pappas, ``{Optimal
  vaccine allocation to control epidemic outbreaks in arbitrary networks},'' in
  \emph{Proceedings of the IEEE Conference on Decision and Control}, 2013, pp.
  7486--7491.

\bibitem{Giamberardino2017}
P.~{Di Giamberardino} and D.~Iacoviello, ``{Optimal control of SIR epidemic
  model with state dependent switching cost index},'' \emph{Biomedical Signal
  Processing and Control}, vol.~31, pp. 377--380, 2017.

\bibitem{Bloem2008}
M.~Bloem, T.~Alpcan, and T.~Basar, ``{Optimal and robust epidemic response for
  multiple networks},'' \emph{Control Engineering Practice}, vol.~17, no.~5,
  pp. 525--533, 2009.

\bibitem{Khanafer}
A.~Khanafer and T.~Basar, ``{An Optimal Control Problem Over Infected
  Networks},'' \emph{Proceedings of the International Conference of Control,
  Dynamic Systems, and Robotics}, no. 125, pp. 1--6, 2014.

\bibitem{Lindmark}
G.~Lindmark and C.~Altafini, ``{Minimum energy control for complex networks},''
  \emph{Scientific Reports}, vol.~8, no.~1, 2018.

\bibitem{Dhingra2018}
N.~K. Dhingra, M.~Colombino, and M.~R. Jovanovic, ``Structured decentralized
  control of positive systems with applications to combination drug therapy and
  leader selection in directed networks,'' \emph{IEEE Transactions on Control
  of Network Systems}, pp. 1--10, 2018.

\bibitem{Torres2017}
J.~A. Torres, S.~Roy, and Y.~Wan, ``{Sparse resource allocation for linear
  network spread dynamics},'' \emph{IEEE Transactions on Automatic Control},
  vol.~62, no.~4, pp. 1714--1728, 2017.

\bibitem{berman1994nonnegative}
A.~Berman and R.~J. Plemmons, \emph{Nonnegative matrices in the mathematical
  sciences}.\hskip 1em plus 0.5em minus 0.4em\relax Siam, 1994, vol.~9.

\bibitem{briat2013robust}
C.~Briat, ``Robust stability and stabilization of uncertain linear positive
  systems via integral linear constraints: L1-gain and l$\infty$-gain
  characterization,'' \emph{International Journal of Robust and Nonlinear
  Control}, vol.~23, no.~17, pp. 1932--1954, 2013.

\bibitem{rantzer2015scalable}
A.~Rantzer, ``Scalable control of positive systems,'' \emph{European Journal of
  Control}, vol.~24, pp. 72--80, 2015.

\bibitem{umenberger2016scalable}
J.~Umenberger and I.~R. Manchester, ``Scalable identification of stable
  positive systems,'' in \emph{Decision and Control (CDC), 2016 IEEE 55th
  Conference on}.\hskip 1em plus 0.5em minus 0.4em\relax IEEE, 2016, pp.
  4630--4635.

\bibitem{Preciado2014}
V.~M. Preciado, M.~Zargham, C.~Enyioha, A.~Jadbabaie, and G.~J. Pappas,
  ``{Optimal resource allocation for network protection against spreading
  processes},'' \emph{IEEE Transactions on Control of Network Systems}, vol.~1,
  no.~1, pp. 99--108, 2014.

\bibitem{Mieghem2011}
P.~Van~Mieghem, ``{The N-intertwined SIS epidemic network model},''
  \emph{Computing}, vol.~93, pp. 147--169, 2011.

\bibitem{Li2012}
C.~Li, R.~{Van De Bovenkamp}, and P.~{Van Mieghem},
  ``{Susceptible-infected-susceptible model: A comparison of N-intertwined and
  heterogeneous mean-field approximations},'' \emph{Physical Review E -
  Statistical, Nonlinear, and Soft Matter Physics}, vol.~86, no.~2, 2012.

\bibitem{dirr2015separable}
G.~Dirr, H.~Ito, A.~Rantzer, and B.~R{\"u}ffer, ``Separable lyapunov functions
  for monotone systems: Constructions and limitations,'' \emph{Discrete Contin.
  Dyn. Syst. Ser. B}, vol.~20, no.~8, pp. 2497--2526, 2015.

\bibitem{manchester2017existence}
I.~R. Manchester and J.-J.~E. Slotine, ``On existence of separable contraction
  metrics for monotone nonlinear systems,'' in \emph{Proceedings of the IFAC
  World Congress}, vol.~50, no.~1, 2017, pp. 8226--8231.

\bibitem{savla2008traveling}
K.~Savla, E.~Frazzoli, and F.~Bullo, ``Traveling salesperson problems for the
  dubins vehicle,'' \emph{IEEE Transactions on Automatic Control}, vol.~53,
  no.~6, pp. 1378--1391, 2008.

\end{thebibliography}
%
%
%

\end{document}